\documentclass[runningheads]{llncs}

\usepackage{eccv}

\usepackage{eccvabbrv}

\usepackage{graphicx}
\usepackage{booktabs}

\usepackage[accsupp]{axessibility}  

\usepackage{hyperref}
\usepackage{makecell}
\usepackage{orcidlink}
\usepackage{multirow}
\usepackage[ruled,linesnumbered]{algorithm2e}
\usepackage{amsmath}
\usepackage{amssymb}
\usepackage[table]{xcolor}
\definecolor{lightgray}{gray}{0.92}

\begin{document}

\title{FutureVLA: Joint Visuomotor Prediction for Vision-Language-Action Model} 


\titlerunning{FutureVLA}

\author{Xiaoxu Xu\inst{1,2} \and
Hao Li\inst{2,3}$^{*}$ \and
Jinhui Ye\inst{2} \and
Yilun Chen\inst{2}$^{\dagger}$ \and
Jia Zeng\inst{2}$^{\dagger}$ \and
Xinyi Chen\inst{2} \and
Linning Xu\inst{2} \and
Dahua Lin\inst{4} \and
Weixin Li\inst{1} \and
Jiangmiao Pang\inst{2}$^{\dagger}$
}

\authorrunning{X. Xu et al.}

\institute{Beihang University.\and Shanghai Artificial Intelligence Laboratory.\and University of Science and Technology of China.\and The Chinese University of Hong Kong.
}


\maketitle
\vspace{-0.4cm}
\begin{abstract}
\let\oldthefootnote\thefootnote%
\renewcommand{\thefootnote}{}%
\footnotetext{$^{*}$ Project leader. $^{\dagger}$ Corresponding authors.}%
\let\thefootnote\oldthefootnote%
Predictive foresight is important to intelligent embodied age-nts. Since the motor execution of a robot is intrinsically constrained by its visual perception of environmental geometry, effectively anticipating the future requires capturing this tightly coupled visuomotor interplay. While recent vision-language-action models attempt to incorporate future guidance, they struggle with this joint modeling. Existing explicit methods divert capacity to task-irrelevant visual details, whereas implicit methods relying on sparse frame pairs disrupt temporal continuity. By heavily relying on visual reconstruction, these methods become visually dominated, entangling static scene context with dynamic action intent. We argue that effective joint visuomotor predictive modeling requires both temporal continuity and visually-conditioned supervision decoupling. To this end, we propose \textbf{FutureVLA}, featuring a novel Joint Visuomotor Predictive Architecture. FutureVLA is designed to extract joint visuomotor embeddings by first decoupling visual and motor information, and then jointly encoding generalized physical priors. Specifically, in the pretraining stage, we leverage heterogeneous manipulation datasets and introduce a Joint Visuomotor Gating mechanism to structurally separate visual state preservation from temporal action modeling. It allows the motor stream to focus on continuous physical dynamics while explicitly querying visual tokens for environmental constraints, yielding highly generalizable joint visuomotor embeddings. Subsequently, in the post-training stage, we employ a latent embeddings alignment strategy, enabling diverse downstream VLA models to internalize these temporal priors without modifying their inference architectures. Our comprehensive evaluation reveals that FutureVLA consistently outperforms baselines without joint visuomotor embeddings guidance, delivering an 11.4\% improvement on the SimplerEnv and a striking 21.7\% gain in real-world robotic operations, validating the effectiveness of our method. \href{https://xuxiaoxxxx.github.io/projects/futurevla/}{Project Website}.

\end{abstract}

\section{Introduction}
An important principle of intelligent control~\cite{lecun2022path} is the ability to anticipate how the world evolves under actions. Rather than reacting solely to current observations, an effective agent should form an internal model of environment dynamics and reason about future transitions. This perspective is closely related to the notion of \emph{world models}~\cite{bruce2024genie,team2025evaluating,assran2023self}, which characterize how states evolve over time conditioned on actions. For vision-language-action (VLA) systems\cite{zheng2024tracevla,li2025robotic}, providing such future guidance is particularly critical: by anticipating an action trajectory's outcome, an agent can better constrain its current decisions. However, robotic actions do not occur in a vacuum; physical motor execution is strictly governed by the visual perception of environmental geometry and spatial affordances. Therefore, fundamentally, this pursuit can be framed as a \textbf{Joint Visuomotor Predictive Modeling} problem: how to effectively capture the intricate interplay between visual environments and motor actions. Motivated by this, recent works~\cite{bu2025univla,zhang2025dreamvla,seer,bruce2024genie, cai2026internvla_a1} leverage future information as supervisory signals.

\begin{figure}[t]
\centering
\includegraphics[width=1.0\linewidth]{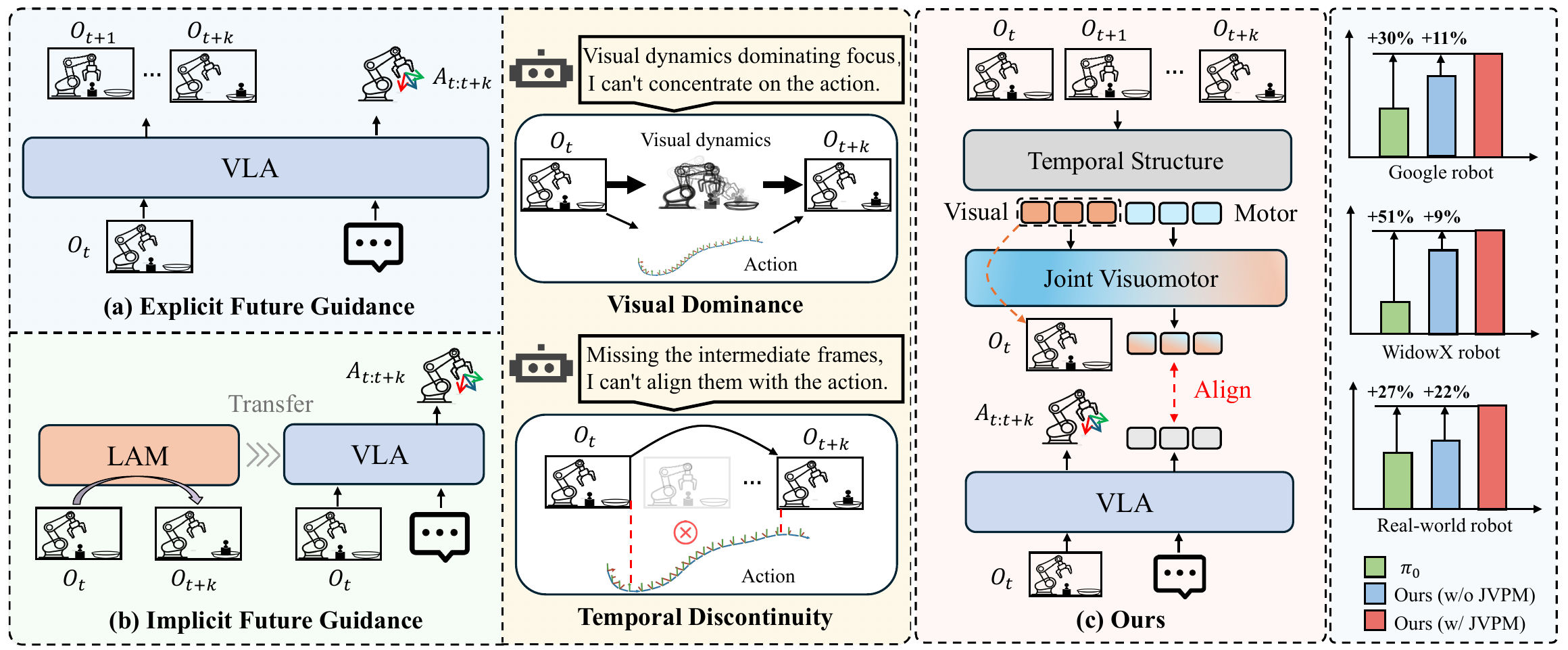} 
\vspace{-0.2cm}
\caption{\small \textbf{Comparison of future guidance paradigms for VLA models.} 
\textbf{(a) Explicit guidance} predicts future video frames. 
\textbf{(b) Implicit guidance} learns latent vectors to reconstruct changes between sparsely sampled frames. 
\textbf{(c) Ours} processes continuous multi-frame clips and structurally decouples the latent representation into a visual stream and a motor stream. This visually-conditioned decoupling extracts joint visuomotor embeddings, yielding consistent performance improvements across diverse benchmarks.} 
\label{fig1}
\vspace{-0.6cm}
\end{figure}

As illustrated in \cref{fig1}, existing paradigms struggle to learn a genuinely visuomotor representation. Explicit future guidance~\cite{worldvla,zhang2025dreamvla,seer,cai2026internvla_a1} predicts future video frames. While intuitive, it inherently prioritizes full-scene reconstruction fidelity over motor logic. This visual dominance diverts substantial capacity to task-irrelevant details. Conversely, implicit future guidance~\cite{ye2025lapa,bu2025univla,bruce2024genie,chen2025villa} attempts to predict latent embeddings between sparsely sampled frames. However, this sparse sampling introduces temporal discontinuity, which disrupts temporal continuity and creates a severe misalignment with the continuous, multi-step nature of robotic action chunks. Moreover, similar to explicit methods, implicit approaches still rely on reconstructing future visual observations. This objective entangles task-irrelevant appearance variations with genuine physical state transitions, causing the learned embeddings to emphasize visual residual changes rather than true motor dynamics.

These limitations reveal a fundamental insight: effective joint visuomotor modeling should respect the distinct physical properties of visual perception and motor execution. Visual embeddings should provide static spatial constraints, while motor embeddings should encapsulate continuous dynamic evolution. Based on these, we propose FutureVLA, a novel framework designed for Joint Visuomotor Predictive Modeling. As shown in \cref{fig1}(c), FutureVLA adopts a concise two-stage paradigm. In the pretraining phase, instead of predicting abstract residual vectors, we directly process continuous multi-frame video clips to extract temporally coherent dynamics that naturally align with the action chunk. Crucially, we structurally decouple these temporal embeddings into two streams: a visual stream dedicated to initial environmental perception, and a motor stream focused purely on physical control. By conditioning the motor stream on the visual stream, we generate physically grounded joint visuomotor embeddings—unified latent representations that seamlessly integrate static environmental constraints with continuous dynamic control intents. In the post-training phase, we use these embeddings as guidance to align with the downstream VLA models.

Specifically, as illustrated in \cref{fig2}, in the Joint Visuomotor Pretraining stage, we leverage heterogeneous manipulation datasets~\cite{open_x_embodiment,simpleenv,Libero} to learn generalized visuomotor priors from continuous multi-frame video clips. To realize the aforementioned decoupling, we design a core Joint Visuomotor Gating mechanism. It divides the temporal representation into two streams with decoupled objectives for visual and motor tokens. Through a gated cross-attention mechanism, the motor stream selectively queries spatial affordances from the visual tokens. This asymmetric interaction prevents visual domination and yields physically grounded joint visuomotor embeddings. In the Joint Visuomotor Embedding Guided VLA Post-training stage, we employ a latent embedding alignment strategy. By aligning the downstream VLA's intermediate representations with these future-aware joint visuomotor embeddings, we enable the VLA to internalize future modeling capabilities.

In summary, our main contributions are threefold:
\vspace{-0.2cm}
\begin{itemize}
    \item We identify two fundamental flaws in existing future guidance methods: visually-dominated embedding entanglement and temporal discontinuity. To address these, we propose \textbf{FutureVLA}, a framework that extracts physically grounded joint visuomotor embeddings to guide downstream VLA models.
    
    \item We introduce a streamlined two-stage training paradigm. During Joint Visuomotor Pretraining, a novel Joint Visuomotor Gating mechanism structurally decouples static visual state preservation from continuous action modeling. In the subsequent post-training stage, a latent alignment strategy transfers these future-aware priors into downstream architectures without altering their inference structures.

    \item Extensive experiments across multiple standard benchmarks demonstrate consistent improvements over strong baselines. With the joint visuomotor embedding guidance, our approach yields absolute average gains of 11.4\% and 9.4\% on SimplerEnv compared to unguided baselines, and surpasses the robust $\pi_0$ by 26.7\% in real-world manipulation tasks.
\end{itemize}
\vspace{-0.6cm}

\section{Related Work}
\label{sec:related}
\noindent\textbf{Vision-Language-Action Models.}
Vision-Language-Action models~\cite{spatialvla,zhen20243d,magma,roboflamingo,huang2025roboground,black2024pi_0,lv2025f1} extend pretrained VLMs to robotic control by generating actions conditioned on visual observations, language instructions, and optional state signals. Existing approaches\cite{RT-1,RT-2,openvla,chen2025internvla} either discretize actions into tokens for autoregressive prediction~\cite{RT-1,RT-2,openvla}, or augment VLM backbones with continuous control heads for direct action regression~\cite{roboflamingo,black2024pi_0,lv2025f1,pi_05,pi_05KI, li2025cronusvla}, typically scaling training on large-scale manipulation datasets\cite{khazatsky2024droid,bu2025agibot,li2026robointer,tian2025interndata,open_x_embodiment}. Despite strong cross-task generalization, most VLA models\cite{li2025switchvla,dong2025interleaved,fan2025interleave} remain largely reactive, predicting actions primarily from current observations without explicitly modeling environment dynamics or future state transitions. Temporal dependencies are usually encoded implicitly within sequence modeling architectures, which limits long-horizon reasoning and the ability to maintain physically consistent action evolution. Recent works introduce auxiliary supervision signals, such as latent embeddings~\cite{ye2025lapa,chen2025villa,bu2025univla,bruce2024genie}, to distill transition dynamics from trajectories, motivating the development of structured temporal representation learning.

\noindent\textbf{Future Guidance in Vision-Language-Action Models.}
Future guidance incorporates predictive signals in VLA systems. Explicit approaches~\cite{seer,worldvla,zhang2025dreamvla,lv2025f1,reconvla} model anticipated future observations or goal states, framing control as visually grounded planning. However, reconstructing full-scene embeddings introduces substantial visual redundancy. Implicit methods~\cite{ye2025lapa,bu2025univla,bruce2024genie,chen2025villa} instead learn latent embeddings from temporally separated frame pairs. Typically, these approaches employ an Inverse Dynamics Model to encode latent embeddings from preceding and subsequent frames, and then use a Forward Dynamics Model to reconstruct the future observation conditioned on the previous frame and inferred action. While this design reduces visual overhead, it often disrupts temporal continuity and entangles visual changes with underlying action dynamics. These limitations underscore the necessity of joint visuomotor predictive modeling, which our framework explicitly addresses. 
\vspace{-0.3cm}

\section{Method}
\begin{figure}[t]
\centering
\includegraphics[width=1.0\linewidth]{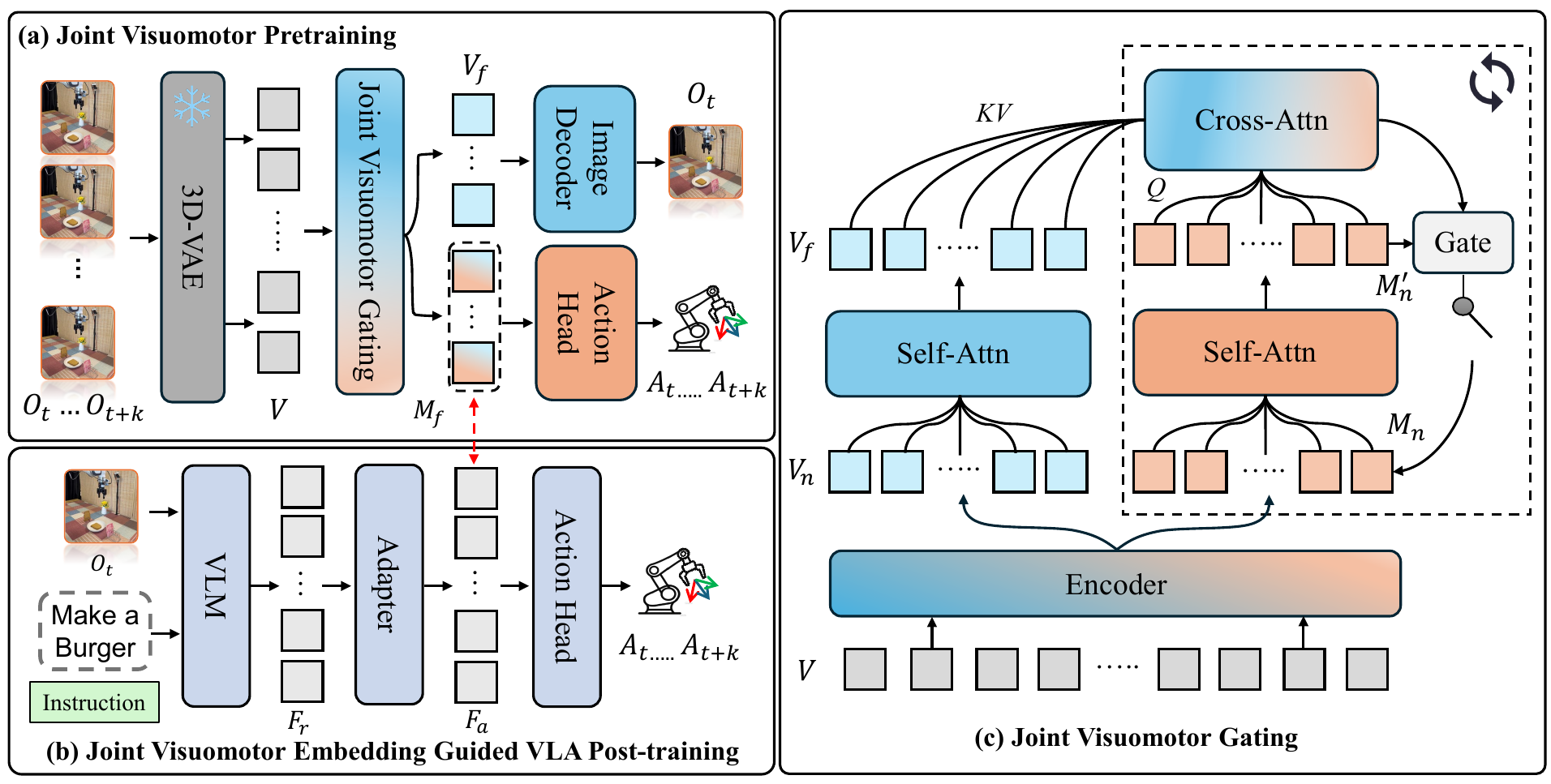} 
\vspace{-0.6cm}
\caption{\small Overview of the FutureVLA framework. 
\textbf{(a) Joint Visuomotor Pretraining:} Continuous video clips are processed by a frozen 3D-VAE into temporal tokens and structurally decoupled into two streams. Visual tokens reconstruct the initial frame $O_t$, while motor tokens, supervised by action chunks, utilize the \textbf{Joint Visuomotor Gating module (c)} based on gated cross-attention, where the motor stream iteratively queries spatial affordances from the visual tokens, yielding physically grounded joint visuomotor embeddings. 
\textbf{(b) Joint Visuomotor Embedding Guided VLA Post-training:} The frozen model provides joint visuomotor embeddings as future-aware temporal priors. Through latent embedding alignment, the downstream VLA's intermediate representations are forced to internalize these dynamics.}
\label{fig2}
\vspace{-0.6cm}
\end{figure}

\vspace{-0.3cm}
As illustrated in \cref{fig2}, we propose FutureVLA, a novel framework designed for Joint Visuomotor Predictive Modeling. FutureVLA operates in two scalable stages: \textbf{(i) Joint Visuomotor Pretraining}, where we pretrain on diverse datasets\cite{open_x_embodiment,simpleenv,Libero}, to efficiently extract physically grounded joint visuomotor embeddings from continuous multi-frame observations via decoupled supervision, and \textbf{(ii) Joint Visuomotor Embedding Guided VLA Post-training}, where these learned embeddings serve as priors to guide downstream VLA through latent embeddings alignment.

\subsection{Joint Visuomotor Embedding Pretraining}
\label{acep}
\noindent\textbf{Visual Tokenization.}
Unlike prior approaches that rely on sparsely sampled frame pairs, as shown in \cref{fig2}(a), we model temporal video clips $\{O_t,\dots,O_{t+k}\}$ to capture richer future-aware context for action prediction. To mitigate the substantial redundancy across adjacent frames, we adopt a pretrained 3D-VAE from WAN~\cite{wan2025wan} to encode input video clips into compact temporal tokens $V$. This step compresses high-frequency visual redundancy while preserving the temporal structure essential for fine-grained action reasoning. During training, the parameters of the 3D-VAE are kept frozen.

\noindent\textbf{Joint Visuomotor Gating.}  
\label{ASCG}
While multi-frame tokens provide rich temporal context, accurate predictive modeling critically depends on visual constraints imposed by the physical environment~\cite{reconvla,chen2025internvla}. Prior works~\cite{chen2025villa,ye2025lapa,bu2025univla} often entangle visual and motor learning into a single reconstruction objective, leading to visually-dominated embeddings that fail to capture pure control logic. To solve this issue, we introduce the Joint Visuomotor Gating mechanism, which achieves visually-conditioned supervision decoupling. 

As shown in \cref{fig2}(c), given the temporal tokens $V$ produced by video tokenization, we first apply an encoder composed of Transformer layers for embedding refinement. The encoded tokens are then partitioned into two subsets with decoupled objectives: visual tokens $V_n$ and motor tokens $M_n$.
Visual tokens $V_n$ are processed via self-attention to capture the global visual context $V_f$. They are supervised exclusively to reconstruct the latent embeddings of the first frame $O_t$, which forces them to preserve informative initial visual information. In parallel, motor tokens $M_n$ are refined through self-attention to obtain $M_n'$. Relieved from the burden of visual rendering, the motor stream can focus purely on continuous physical dynamics. Moreover, to ensure actions are physically plausible within the environment, we apply cross-attention using $M_n'$ as queries and $V_f$ as keys and values. This allows the motor stream to selectively retrieve necessary geometric and spatial constraints from the static visual state.

To model the influence of these visual constraints, we introduce a gating mechanism. A learnable scalar parameter $r$ controls the contribution of the cross-attended embeddings. The gated embeddings are then combined with the original motor tokens to form updated action representations. 

\begin{equation}
\begin{aligned}
M_n' &= \mathrm{SelfAttn}(M_n), \\
M_{n}' &= \mathrm{CrossAttn}(M_n', V_f, V_f), \\
M_{n+1} &= \sigma(r) \odot M_{n}' + M_n',
\end{aligned}
\end{equation}
where $\sigma(\cdot)$ denotes the sigmoid function. After refinement steps, the final joint visuomotor embeddings $M_f$ are obtained.


\noindent\textbf{Visual Reconstruction.}
\label{pvfc}
As shown in \cref{fig2}(a), given the visual representation $V_f$, we reconstruct the first-frame observation $O_t$. Here we use the embedding $V_t$ produced by the pretrained 3D-VAE\cite{wan2025wan} as the reconstruction target, encouraging the model to focus on semantically meaningful information while reducing computational overhead\cite{assran2025v,seer,worldvla}. The image decoder consists of Transformer layers that transform $V_f$ into the reconstructed latent representation $V_r$. The reconstruction loss is defined as:

\begin{equation}
\mathcal{L}_I={\left\|V_r-V_t\right\|}_2\\\\.
\end{equation}

\noindent\textbf{Predict Motor Chunking.}
\label{actionmodel}
To ensure that the proposed joint visuomotor embeddings can be seamlessly integrated into diverse post-training paradigms, we instantiate two representative action heads: an OFT-style model and a GR00T-style model, allowing our FutureVLA to remain agnostic to specific action modeling choices:\\
\textbf{(1) OFT-style.}
Following OpenVLA-OFT\cite{openvla-oft}, we adopt a lightweight ResNet~\cite{he2016identity} composed of two residual MLP layers. The joint visuomotor embeddings $M_f$ are mapped to action predictions $\hat A_{t:t+k}$, which regresses the target action chunk $A_{t:t+k}$ using the MAE loss:
\begin{equation}
\mathcal{L}_A = {\left\|\hat A_{t:t+k}-A_{t:t+k}\right\|}_1\\\\.
\end{equation}
\textbf{(2) GR00T-style.}
For GR00T-style action modeling\cite{bjorck2025gr00t}, we adopt a conditional flow matching formulation. The joint visuomotor embeddings $M_f$ serve as conditioning inputs, while the variable  $X_t$ represents the action chunk $A_{t:t+k}$. We train a neural vector field $v_\tau^\theta$ to minimize the flow matching objective:
\begin{equation}
\mathcal{L}_A = \mathbb{E}_{p(X_t|M_f), q(X_t^\tau|X_t)} \left\| v_\tau^\theta(X_t^\tau, M_f) - u(X_t^\tau \mid X_t) \right\|_2,
\end{equation}
where $\tau \in [0, 1]$ denotes the flow-matching time variable. In practice, we sample Gaussian noise $\epsilon \sim N(0, I)$ and construct the noisy action as
\begin{equation}
X_t^\tau = \tau X_t + (1 - \tau)\epsilon.
\end{equation}
The target vector field is defined as $u(X_t^\tau \mid X_t) = \epsilon - X_t$. and the network $v_\tau^\theta$ is trained to predict this denoising direction. During training, $\tau$ is sampled from a beta distribution following standard practice in conditional flow matching.
The overall joint visuomotor pretraining objective combines reconstruction and action supervision:
\begin{equation}
\mathcal{L}_1 = \lambda \mathcal{L}_I + \mathcal{L}_A,
\end{equation}
where $\lambda$ balances the reconstruction loss and the action loss.

\subsection{Joint Visuomotor Embedding Guided VLA Post-training}
\label{PS}

After obtaining the joint visuomotor embeddings, the next challenge is effectively transferring these temporal priors to downstream VLA models. Most existing implicit approaches~\cite{ye2025lapa,bu2025univla} rely on latent guidance merely through pretrained weight initialization.  Motivated by \cite{wei2022chain,chen2025villa,li2025spatial}, we utilize a latent embedding alignment strategy. This approach treats our extracted joint visuomotor embeddings $M_f$ as semantic anchors, forcing the intermediate representations of the VLA network to maintain consistency with physically grounded future dynamics without requiring multi-frame inputs during inference. Furthermore, prior works\cite{ye2025lapa,bu2025univla,chen2025villa} typically validates extracted latent embeddings guidance within a single downstream architecture. 

As illustrated in \cref{fig2}(b), during the post-training phase, we freeze the pretrained weights from the pretraining stage to extract high-fidelity joint visuomotor embeddings $M_f$ from future video clips. Concurrently, the current observation $O_t$ and instruction are fed into a vision-language model backbone~\cite{yang2025qwen3}, producing intermediate representations $F_r$. 
To bridge the representational gap, we introduce a lightweight Transformer adapter between the VLA representations and the joint visuomotor embeddings. The adapted embeddings $F_a$ are aligned with $M_f$ via an MSE-based alignment loss. The overall post-training objective is:
\begin{equation}
\mathcal{L}_2 = \beta \| M_f - F_a \|_2 + \mathcal{L}_A
\end{equation}
where $\beta$ balances the latent alignment loss and the action loss $\mathcal{L}_A$.

\section{Experiments}
\subsection{Experimental Setups}
We evaluate our method on both simulation benchmarks and real-world robotic tasks. 
In simulation, we adopt two widely used VLA benchmarks: LIBERO\cite{Libero} and SimplerEnv\cite{simpleenv}. 
For LIBERO, following prior work\cite{instructvla,reconvla,li2025cronusvla}, we evaluate on four standard task suites: \textit{Object}, \textit{Spatial}, \textit{Goal} and \textit{Long}, which collectively assess language grounding, compositional reasoning, and long-horizon manipulation under a unified robot embodiment. For SimplerEnv, we evaluate on both the Google robot and WidowX robot task suites, which encompass diverse manipulation scenarios across different robot embodiments. In the real world, we further evaluate our method on four manipulation tasks using a Franka robot: \textit{make a burger}, \textit{insert roses into a pot}, \textit{scoop beans with a spoon}, and \textit{erase handwriting on a whiteboard}. 
These tasks cover a broad spectrum of manipulation primitives, including grasping, tool use, insertion, and contact-rich control. We compare our method with representative VLA approaches that span three
categories: original VLA baselines, explicit future guidance methods, and implicit
future guidance methods. For more detailed information, please see the Appendix.

\begin{table}[t]
\centering
\caption{Comparison of different VLA models across four tasks in two SimplerEnv settings on the Google robot. ``GT'' indicates GR00T-style model, ``OT'' denotes OFT-style model.}

\setlength{\tabcolsep}{0.2mm}{
\begin{tabular}{ccccccc}
\toprule
Settings & Methods
& \makecell{\small Pick\\Coke Can}
& \makecell{\small Move\\Near}
& \makecell{\small Open/Close\\Drawer}
& \makecell{\small Put in\\Drawer} & Avg. \\
\midrule

\multirow{6}{*}{\makecell{\small Visual\\Matching}}
& OpenVLA-OFT\cite{openvla-oft}
& 72.3 & 69.6 & 47.2 & 0.9  & 47.5 \\
& $\pi_0$\cite{black2024pi_0}
& 87.3 & 35.0 & \textbf{72.6} & 16.0 & 52.7 \\
& GR00T-N1.5\cite{bjorck2025gr00t}
& 51.7 & 54.0 & 27.8 & 7.4  & 35.2 \\
& Villa-X\cite{chen2025villa}
& \textbf{98.7} & 75.0 & 59.3 & 5.6  & 59.6 \\
& \cellcolor{lightgray}FutureVLA-GT
& \cellcolor{lightgray}92.3 & \cellcolor{lightgray}74.2
& \cellcolor{lightgray}68.5 & \cellcolor{lightgray}\textbf{85.2}
& \cellcolor{lightgray}\textbf{80.1} \\
& \cellcolor{lightgray}FutureVLA-OT
& \cellcolor{lightgray}97.0 & \cellcolor{lightgray}\textbf{83.8}
& \cellcolor{lightgray}55.6 & \cellcolor{lightgray}74.1
& \cellcolor{lightgray}77.6 \\

\midrule

\multirow{5}{*}{\makecell{\small Variant\\Aggregation}}
& OpenVLA-OFT\cite{openvla-oft}
& 65.3 & 59.0 & 12.2 & 0.5  & 34.3 \\
& $\pi_0$\cite{black2024pi_0}
& 85.2 & 40.8 & 42.1 & 16.0 & 46.0 \\
& GR00T-N1.5\cite{bjorck2025gr00t}
& 69.3 & 68.7 & 35.8 & 4.0  & 44.5 \\
& \cellcolor{lightgray}FutureVLA-GT
& \cellcolor{lightgray}\textbf{95.7} & \cellcolor{lightgray}\textbf{72.9}
& \cellcolor{lightgray}\textbf{52.4} & \cellcolor{lightgray}\textbf{81.2}
& \cellcolor{lightgray}\textbf{75.6} \\
& \cellcolor{lightgray}FutureVLA-OT
& \cellcolor{lightgray}80.5 & \cellcolor{lightgray}61.5
& \cellcolor{lightgray}31.7 & \cellcolor{lightgray}23.8
& \cellcolor{lightgray}49.4 \\

\bottomrule
\end{tabular}
}
\label{Tab.1}
\end{table}
\vspace{-0.4cm}

\begin{table}[t]
    \centering
    \caption{Comparison of different VLA models across four tasks in the SimplerEnv setting on the WidowX robot. }    
    \setlength{\tabcolsep}{0.8mm}{
    \begin{tabular}{cccccc}
        \toprule
        Methods & \makecell{\small Put \\ Spoon}
        & \makecell{\small Put \\Carrot}
        & \makecell{\small Stack \\Cube}
        & \makecell{\small Put \\Eggplant}
        & Avg. \\
        \midrule
        OpenVLA-OFT\cite{openvla-oft}   & 12.5 & 4.2  & 4.2  & \textbf{100.0} & 30.2 \\
        $\pi_0$\cite{black2024pi_0}      & 25.0 & 16.7 & 12.5 & 29.2  & 20.9 \\
        GR00T-N1.5\cite{bjorck2025gr00t}   & 75.3 & 54.3 & \textbf{57.0} & 61.3  & 61.9 \\
        UniVLA\cite{bu2025univla}       & 52.8 & 55.6 & 2.8  & 80.6  & 47.9 \\
        Villa-X\cite{chen2025villa}      & 48.3 & 24.2 & 19.2 & 71.7  & 40.8 \\

        \cellcolor{lightgray}FutureVLA-GT
        & \cellcolor{lightgray}\textbf{83.3}
        & \cellcolor{lightgray}\textbf{75.0}
        & \cellcolor{lightgray}29.2
        & \cellcolor{lightgray}\textbf{100.0}
        & \cellcolor{lightgray}\textbf{71.9} \\

        \cellcolor{lightgray}FutureVLA-OT
        & \cellcolor{lightgray}75.0
        & \cellcolor{lightgray}54.2
        & \cellcolor{lightgray}25.0
        & \cellcolor{lightgray}\textbf{100.0}
        & \cellcolor{lightgray}63.6 \\

        \bottomrule
    \end{tabular}
}
\label{Tab.2}
\vspace{-0.2cm}
\end{table}

\begin{table}[t]
    \centering
    \caption{Comparison of different VLA models on the LIBERO benchmark. }
    \setlength{\tabcolsep}{2mm}
    \begin{tabular}{cccccc}
        \toprule
        Methods & Object & Spatial & Goal & Long & Avg. \\
        \midrule
        OpenVLA-OFT\cite{openvla-oft}   & 88.4 & 84.7 & 79.2 & 53.7 & 76.5 \\
        $\pi_0$\cite{black2024pi_0}       & 98.8 & 96.8 & 95.8 & 85.2 & 94.2 \\
        GR00T-N1.5\cite{bjorck2025gr00t}    & 97.6 & 94.4 & 93.0 & 90.6 & 93.9 \\
        WorldVLA\cite{worldvla}             & 96.2 & 87.6 &  83.4 & 60.0 & 81.8 \\
        DreamVLA\cite{zhang2025dreamvla}      & 94.0 & 97.5 & 89.5 & 89.5 & 92.6 \\
        UniVLA\cite{bu2025univla}        & 96.8 & 96.5 & 95.6 & 92.0 & 95.2 \\
        Villa-X\cite{chen2025villa}       & 97.0 & 97.5 & 91.5 & 74.5 & 90.1 \\
        \cellcolor{lightgray}FutureVLA-GT
        & \cellcolor{lightgray}\textbf{99.8}
        & \cellcolor{lightgray}98.8
        & \cellcolor{lightgray}98.6
        & \cellcolor{lightgray}\textbf{96.0}
        & \cellcolor{lightgray}\textbf{98.3} \\

        \cellcolor{lightgray}FutureVLA-OT
        & \cellcolor{lightgray}99.2
        & \cellcolor{lightgray}\textbf{99.6}
        & \cellcolor{lightgray}\textbf{99.2}
        & \cellcolor{lightgray}94.8
        & \cellcolor{lightgray}98.2 \\

        \bottomrule
    \end{tabular}
\label{Tab.3}
\vspace{-0.4cm}
\end{table}

\begin{figure}[!t]
  \centering
   \includegraphics[width=1\linewidth]{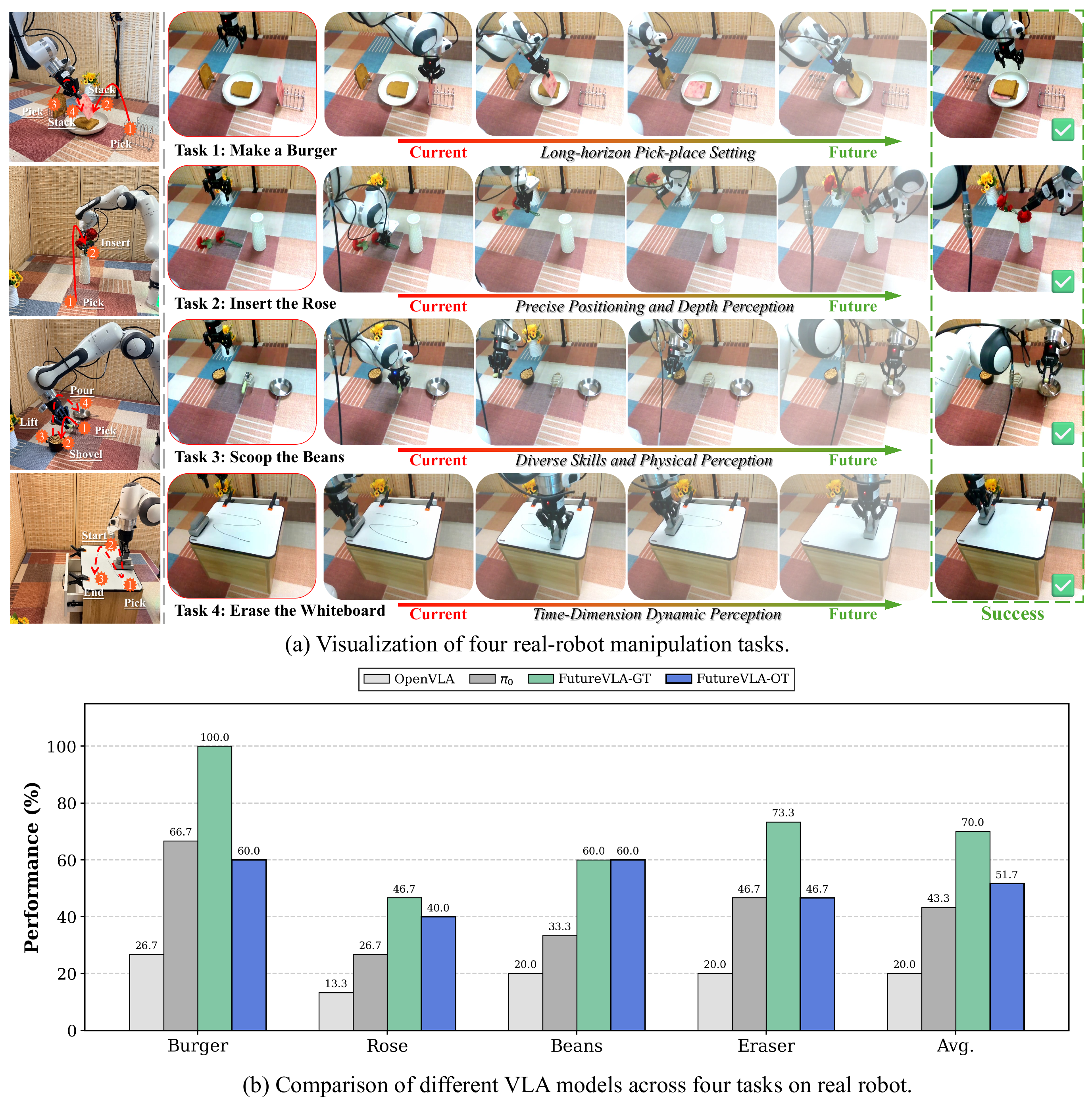}
   \vspace{-0.6cm}
   \caption{\small Visualization in (a) and performance comparison in (b) for four real-robot manipulation tasks.
   In (a), from top to bottom: 
   (1) \textit{Make a burger}: The robot first picks up the burger slice on the right and places it on the bread, then picks up the bread slice on the left and places it on top of the burger slice. (2) \textit{Insert roses into a pot}: The robot picks up the roses from the table and inserts them into the pot. (3) \textit{Scoop beans with a spoon}: The robot grasps the spoon in the center, scoops beans from the left container, and transfers them into the bowl on the right. (4) \textit{Erase handwriting on a whiteboard}: The robot erases the black markings on the whiteboard.}
   \label{fig3}
   \vspace{-0.6cm}
\end{figure}

\subsection{Main Results}
\noindent\textbf{Real-to-sim Evaluation on SimplerEnv.}
We evaluate FutureVLA on the SimplerEnv\cite{simpleenv} benchmark using both Google robot and WidowX robot task suites.  As shown in \cref{Tab.1}, on the Google robot tasks, FutureVLA consistently achieves superior performance across both GR00T-style and OFT-style frameworks. Compared to strong baselines, our method yields remarkable improvements. Under the Visual Matching setting, FutureVLA-GT achieves an absolute average gain of 44.9\% over the GR00T-N1.5\cite{bjorck2025gr00t}, while FutureVLA-OT improves upon OpenVLA-OFT\cite{openvla-oft} by 30.1\%. Notably, our method yields particularly large gains on long-horizon tasks such as \textit{Put in Drawer}. Similar trends are observed on the WidowX robot, as demonstrated in \cref{Tab.2}, demonstrating that our FutureVLA provides temporally coherent and physically grounded guidance.

\noindent\textbf{Simulated Evaluation on LIBERO.}
We further report results on the LIBERO benchmark, as shown in \cref{Tab.3}. Introducing joint visuomotor embeddings leads to consistent performance improvements across all tasks. The improvement is most pronounced on the \textit{Long} tasks, indicating that modeling extensive temporal contexts enables the VLA model to better capture long-range dependencies and future state evolution. 

\noindent\textbf{Real-world Evaluation.}
We validate our method on a real robot. As shown in \cref{fig3}, across four complex manipulation tasks, FutureVLA consistently outperforms existing generalist policies. FutureVLA-GT achieves an average success rate of 70.0\%, vastly surpassing the robust $\pi_0$ by 26.7\%. Notably, the improvement is particularly pronounced for tasks requiring fine-grained and continuous control, such as the contact-rich whiteboard erasing task. These findings highlight the clear advantage of joint visuomotor predictive modeling for sustained, real-world physical execution.

\vspace{-0.4cm}
\subsection{Ablation Studies}

\noindent\textbf{Effectiveness of Joint Visuomotor Predictive Modeling.}
To explicitly validate the contribution of the temporal priors extracted during pretraining, we conduct an ablation study isolating the  Joint Visuomotor Predictive Modeling (JVPM) guidance during the post-training stage. As shown in \cref{Tab.4}, integrating JVPM yields substantial performance improvements on WidowX robot across both VLA architectures. Specifically, aligning the VLA's intermediate representations with our joint visuomotor embeddings brings an absolute average gain of 9.4\% for both the GR00T-style and OFT-style models compared to its non-guided counterpart (wo/ JVPM). This demonstrates that our joint visuomotor design not only leverages multi-frame priors as complementary supervision, but also structurally decouples and conditions visual and motor representations to capture physically grounded future dynamics. These learned dynamics can be seamlessly transferred to downstream models, enhancing reasoning capabilities. 

\begin{table}[!t]
    \centering
    \caption{Ablation of Joint Visuomotor Predictive Modeling guidance on the WidowX robot. ``w/ JVPM'' and ``wo/ JVPM'' denote whether latent embedding alignment is applied during post-training.} 
    \setlength{\tabcolsep}{0.8mm}
    \begin{tabular}{cccccc}
        \toprule
        Methods & \makecell{\small Put \\ Spoon}
        & \makecell{\small Put \\Carrot}
        & \makecell{\small Stack \\Cube}
        & \makecell{\small Put \\Eggplant}
        & Avg. \\
        \midrule
        FutureVLA-GT wo/ JVPM & 79.2 & 58.3 & 20.8 & 91.7 & 62.5 \\
        FutureVLA-GT w/ JVPM
        & \textbf{83.3} & \textbf{75.0} & \textbf{29.2} & \textbf{100.0} & \textbf{71.9} \\
        \midrule
        FutureVLA-OT wo/ JVPM & 66.7 & 58.3 & 12.5 & 79.2 & 54.2 \\
        FutureVLA-OT w/ JVPM
        & \textbf{75.0} & \textbf{54.2} & \textbf{25.0} & \textbf{100.0} & \textbf{63.6} \\
        \bottomrule
    \end{tabular}
\label{Tab.4}
\end{table}

\begin{table*}[!t]
    \centering
    \caption{Ablation studies of Joint Visuomotor Gate on the WidowX robot. ``MT'' represents motor tokens, ``VT'' indicates visual tokens and ``JVG'' denotes Joint Visuomotor Gating.}
    \setlength{\tabcolsep}{1.5mm}
    \begin{tabular}{c c c c|ccccc}
        \toprule
        & MT & VT & JVG 
        & \makecell{\small Put \\ Spoon}
        & \makecell{\small Put \\Carrot}
        & \makecell{\small Stack \\Cube}
        & \makecell{\small Put \\Eggplant} & Avg. \\
        \midrule
        (a) &  &  & 
        & 79.2 & 58.3 & 20.8 & 91.7 & 62.5 \\

        (b) & \checkmark &  & 
        & 66.7 & 54.2 & 16.7 & 95.8 & 58.4 \\

        (c) & \checkmark & \checkmark & 
        & 66.7 & 62.5 & 37.5 & 95.8 & 65.6 \\

        (d) & \checkmark & \checkmark & \checkmark
        & \textbf{83.3} & \textbf{75.0} & \textbf{29.2} & \textbf{100.0} & \textbf{71.9} \\
        \bottomrule
    \end{tabular}
    \label{Tab.5}
\vspace{-0.6cm}
\end{table*}

\noindent\textbf{Ablation on Joint Visuomotor Gate.}
To analyze the contribution of the Joint Visuomotor Gate, we conduct ablation studies on the WidowX robot using the GR00T-style framework as the base architecture. The results are reported in \cref{Tab.5}. Variant (a) corresponds to the vanilla GR00T-style baseline trained solely with action supervision, without any future guidance. Variant (b) introduces temporal continuity by processing multi-frame video tokens directly through Transformer layers to produce motor tokens, which are then fed into the action head. Then these pretrained motor tokens are utilized to guide downstream VLA model. Notably, this variant results in a slight performance degradation compared to the baseline. This aligns with our theoretical analysis: while multi-frame inputs provide future context, directly mapping them to actions without explicit visual-aware constraints forces the network to absorb noisy or task-irrelevant visual dynamics, failing to provide reliable physical guidance. Variant (c) introduces decoupled supervision. The video tokens are explicitly decomposed into visual tokens and motor tokens, but no interaction between the two streams is enabled. The VT stream is supervised by the image decoder to reconstruct the first-frame latent embeddings, while the MT stream predicts the action chunk. This structural decoupling forces the network to separate static visual information from dynamic motion, preventing the motor embeddings from degenerating into visual residuals. This decoupling alone results in a 7.2\% performance boost over Variant (b), demonstrating the critical importance of isolating visual rendering from action modeling. Finally, Variant (d) represents our full FutureVLA framework. We inject the static visual information from VT into the dynamic MT stream via the proposed Joint Visuomotor Gating. This mechanism allows the motor tokens to explicitly query the visual tokens for environmental constraints  before predicting the action. This visually-conditioned supervision decoupling further improves the overall task success rate by 65.6\% to 71.9\%. This result compellingly confirms that selectively conditioning dynamic action modeling on static environmental priors is essential for extracting pure, physically grounded joint visuomotor embeddings.

\begin{figure}[!t]
    \centering
    \begin{minipage}{0.48\linewidth}
        \centering
        \includegraphics[width=\linewidth]{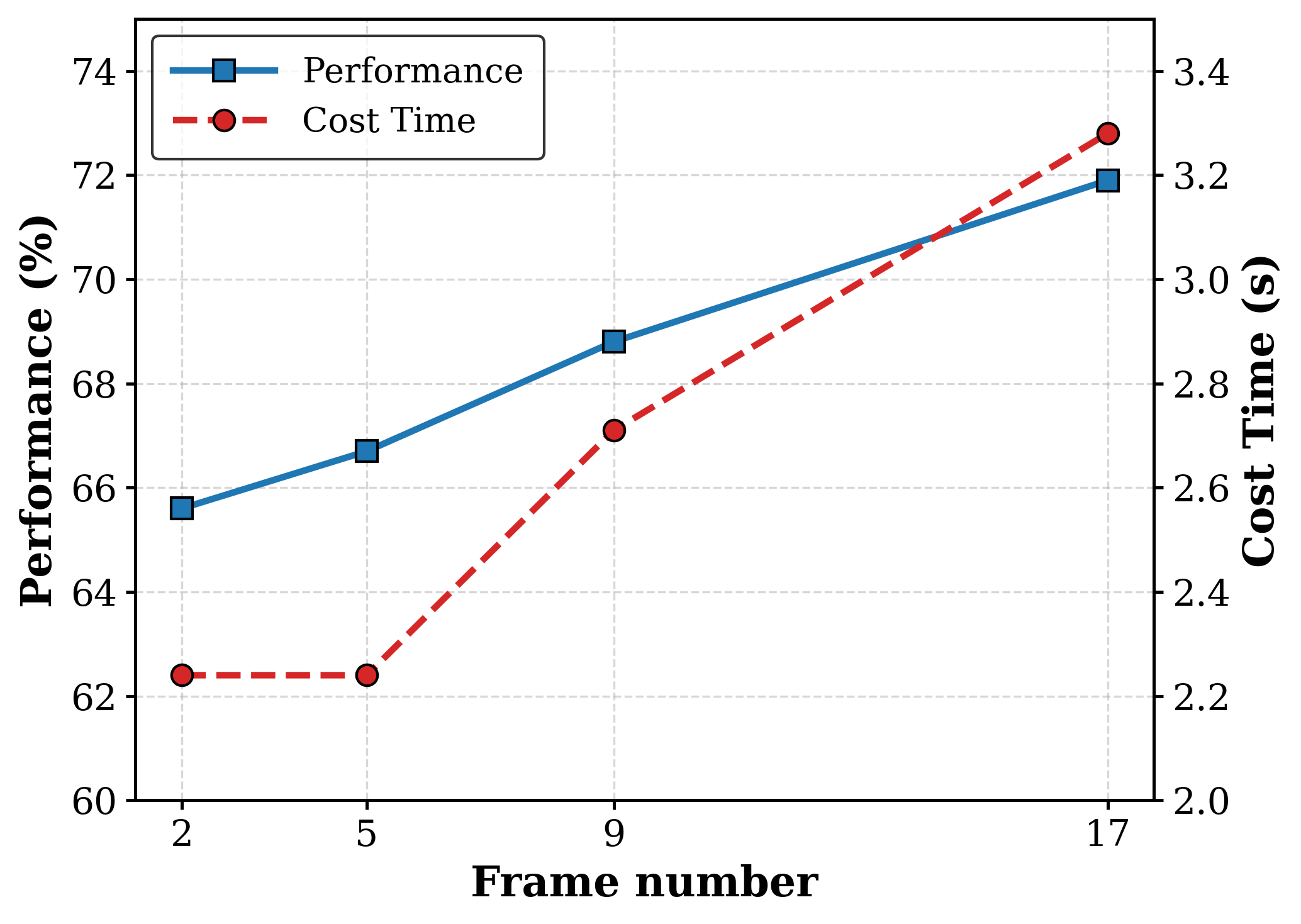}
        \caption{\small Effect of different temporal sampling density on joint visuomotor learning, based on the WidowX robot.}
        \label{fig4}
    \end{minipage}
    \hfill
    \begin{minipage}{0.48\linewidth}
        \centering
        \includegraphics[width=\linewidth]{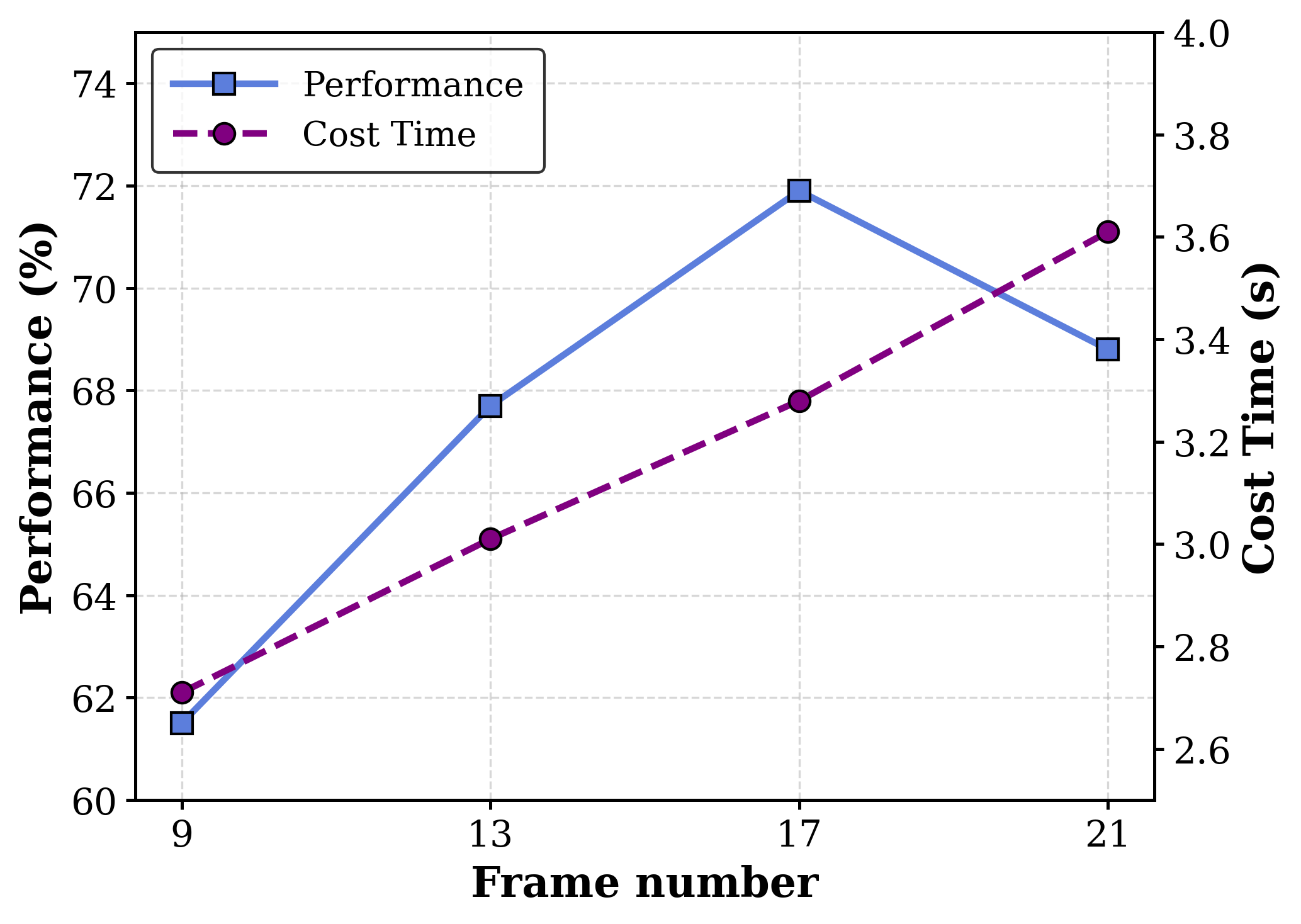}
        \caption{\small Performance of joint visuomotor learning with varying temporal horizon length on the WidowX robot.}
        \label{fig5}
    \end{minipage}
\vspace{-0.3cm}
\end{figure}

\begin{figure}[t]
  \centering
   \includegraphics[width=0.9\linewidth]{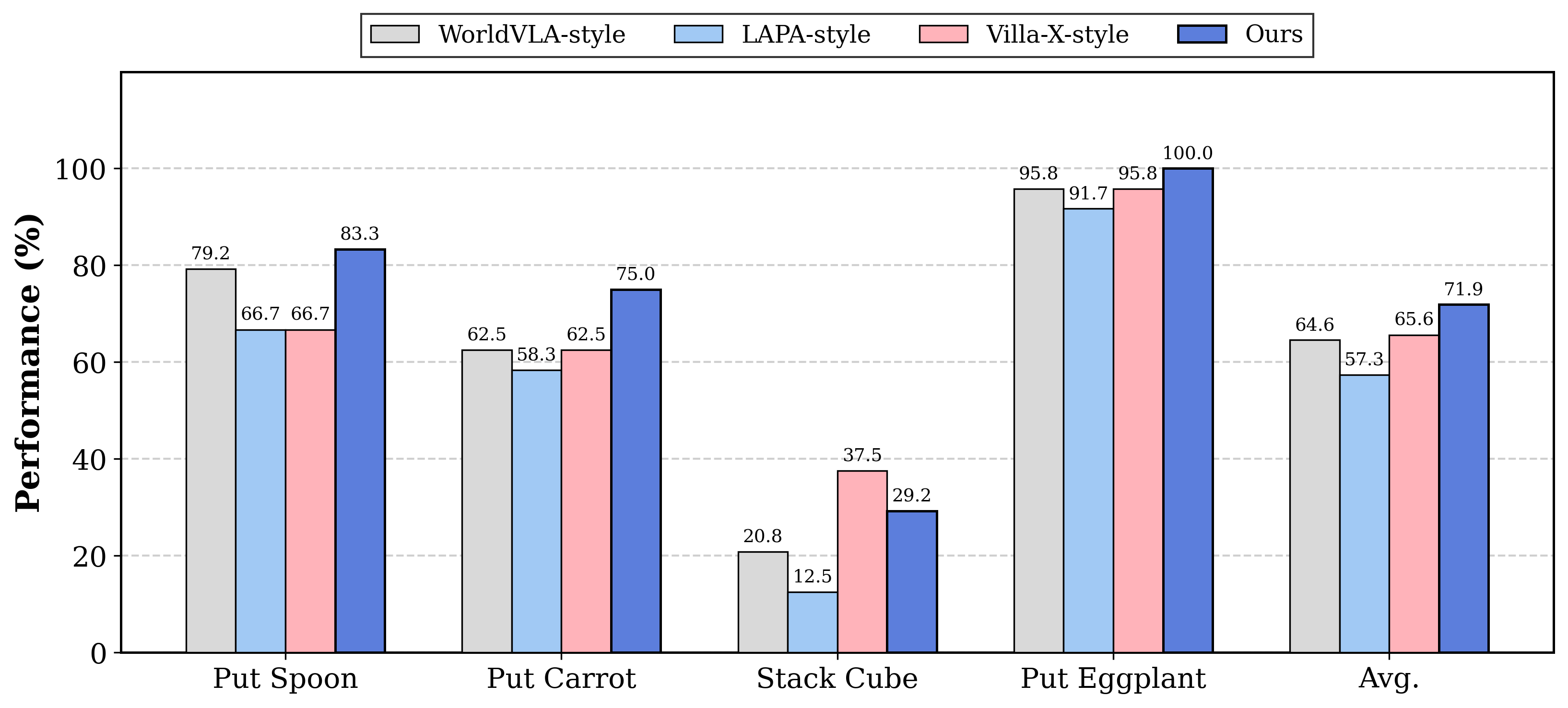}
   \vspace{-0.4cm}
   \caption{\small Performance comparison of different future guidance styles on the WidowX robot under a unified training framework. Explicit future guidance is evaluated using the WorldVLA-style, while implicit future guidance is evaluated using two representative paradigms: LAPA-style and Villa-X-style.}
   \label{fig6}
   \vspace{-0.6cm}
\end{figure}

\begin{figure}[t]
  \centering
   \includegraphics[width=0.8\linewidth]{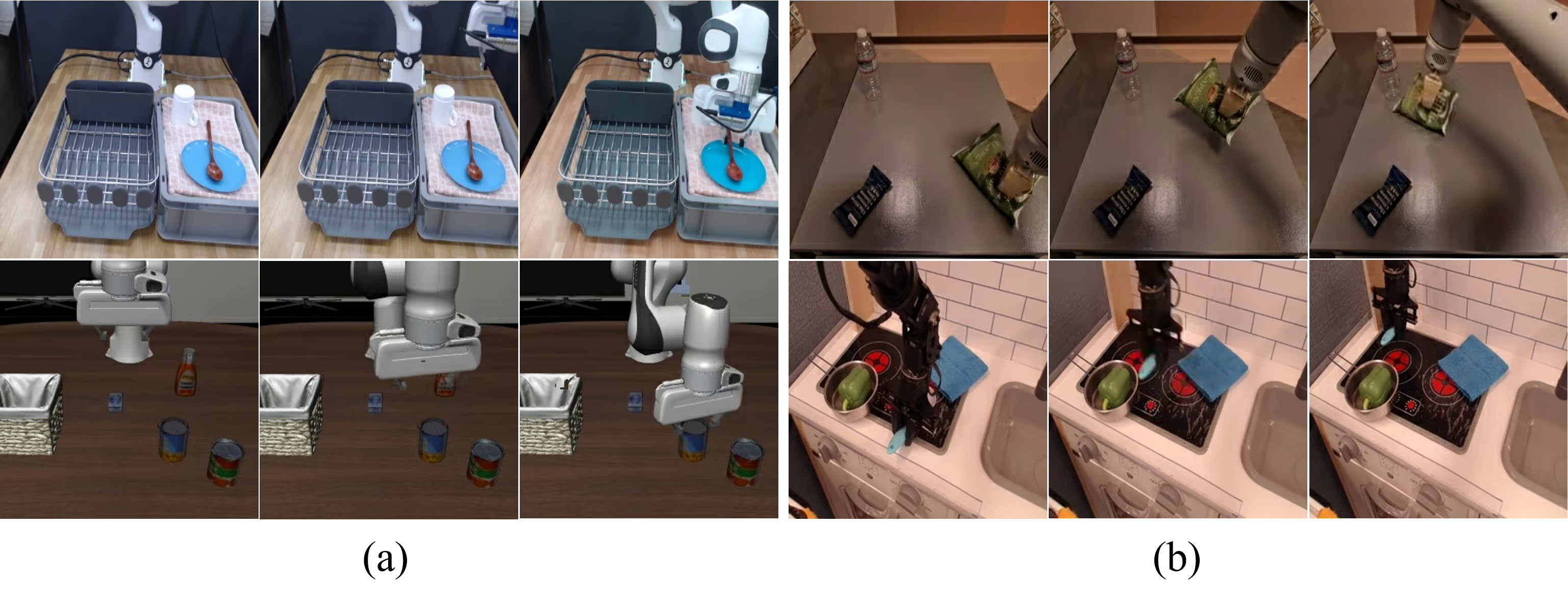}
   \vspace{-0.4cm}
   \caption{\small Visualization of video clips associated with similar latent embeddings. In (a), both clips exhibit a grasping action followed by a front-to-back translation and downward movement. In (b), both clips show a grasping action with a upper-left translation.}
   \label{fig7}
   \vspace{-0.3cm}
\end{figure}


\begin{figure}[t]
  \centering
   \includegraphics[width=0.7\linewidth]{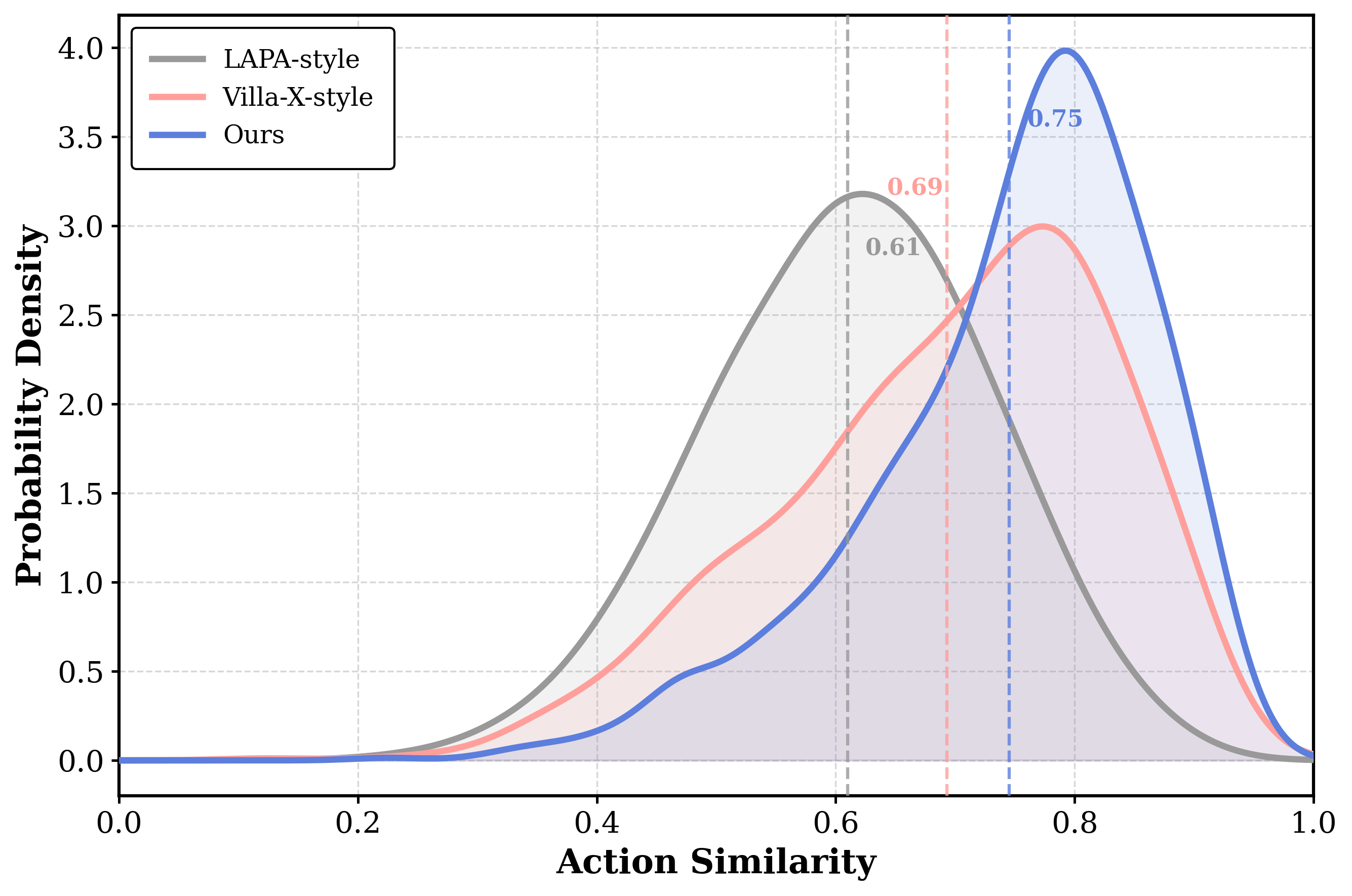}
   \caption{\small Probability density distributions of LAPA-style, Villa-X-style, and our method over the similarity between latent embeddings and executed action chunks. Vertical lines denote the corresponding distribution means.}
   \label{fig8}
   \vspace{-0.7cm}
\end{figure}

\noindent\textbf{Necessity of Continuous Temporal Alignment.}
A core claim of our framework is that continuous temporal modeling is superior to the sparse sampling paradigms used in existing implicit methods. To validate this, we analyze the contribution of inter-frame temporal density by varying the number of sampled frames from a fixed-length video clip during the Joint Visuomotor Pretraining stage. Starting from a 17-frame horizon, we retain the first and last frames and uniformly sample intermediate frames, resulting in sampling sizes of $\{2, 5, 9, 17\}$, as illustrated in~\cref{fig4}. Due to the architectural constraints of the 3D-VAE~\cite{wan2025wan}, which requires inputs in a $4N+1$ format, the two-frame configuration is padded to five frames via repetition. The results demonstrate a clear and consistent performance degradation as temporal sparsity increases. This strongly confirms that relying on sparse frame pairs (e.g., the 2-frame or 5-frame settings) disrupts temporal continuity, whereas leveraging continuous multi-frame observations substantially enhances physically grounded action modeling.

\noindent\textbf{Temporal Horizon and Action Alignment.}
While continuous frames are essential, the length of the temporal horizon should also be carefully calibrated. We conduct a representative study on WidowX robot, varying the input video sequence length among $\{9,13,17,21\}$ frames, with results shown in~\cref{fig5}. As the sequence length increases from 9 to 17, performance steadily improves, indicating that a longer temporal context provides richer future-aware cues for joint visuomotor predictive modeling. However, extending the horizon to 21 frames causes performance to degrade. This is primarily due to the fixed action chunking horizon of 16. Too few frames lead to insufficient temporal coverage for the predicted actions, while overly long sequences introduce redundant or temporally misaligned context. Consequently, we select 17 frames as the optimal default horizon to perfectly align the visual observation length with the motor execution length.

\noindent\textbf{Fair Comparison under a Unified Architecture.}
In previous comparisons, different future guidance methods often rely on heterogeneous backbone architectures, which may introduce confounding factors in performance evaluation. To enable a fair and controlled comparison, we re-implement three representative future guidance methods within a unified architecture. Specifically, we adopt the implementation protocol of WorldVLA\cite{worldvla} for explicit future guidance, and follow the LAPA\cite{ye2025lapa} and Villa-X\cite{chen2025villa} protocols for implicit future guidance. As shown in \cref{fig6}, although adopting a stronger backbone consistently improves the performance of existing future guidance methods, our approach achieves the best results under identical architectural. These results indicate that existing explicit or implicit methods rely on predicting future visual states, forcing the latent space to entangle visual rendering with motor control. In contrast, our decoupled approach isolates pure motor intent, leading to more robust guidance.

\noindent\textbf{Physical Consistency of Joint Visuomotor Embeddings.}
A core claim of our framework is that it captures true motor intent rather than mere visual residuals. As qualitatively shown in~\cref{fig7}, video clips corresponding to nearby latent embeddings in our space exhibit highly consistent robot behaviors, suggesting that latent proximity deeply reflects similarity in low-level physical execution. To quantitatively assess this property, we compare LAPA-style\cite{ye2025lapa}, Villa-X-style\cite{ye2025lapa}, and our FutureVLA by analyzing the alignment between latent embedding similarity and executed action consistency. Specifically, we compute cosine similarities between latent embeddings. For the corresponding action chunks, we measure their behavioral consistency using a proposed Physics-Aware Action Consistency metric (see Appendix). For each target embedding, we retrieve its top-$3$ nearest neighbors based on embedding similarity and extract the corresponding action consistency scores between these retrieved pairs. The probability density distributions in~\cref{fig8} demonstrate that our FutureVLA exhibits the strongest correlation between latent similarity and physical action consistency among all methods. This result compellingly indicates that our joint visuomotor predictive modeling faithfully captures underlying physical dynamics, avoiding the pitfall of degenerating into visual differences.

\vspace{-0.3cm}
\section{Conclusion}
\vspace{-0.3cm}
In this work, we address the visually-dominated embedding entanglement and temporal discontinuity in existing future-guided VLAs by proposing \textbf{FutureVLA}. By structurally decoupling static visual constraints from continuous temporal action modeling, FutureVLA extracts physically grounded joint visuomotor embeddings. Through a streamlined two-stage paradigm of pretraining and latent alignment post-training, we efficiently transfer these future-aware priors into diverse VLA policies without altering their inference architectures. Extensive experiments across simulated and real-world benchmarks demonstrate that FutureVLA significantly improves reasoning and execution capabilities over other future guidance methods. By isolating true motor intent from superficial visual variations, FutureVLA offers a scalable path toward physically consistent embodied foundation models.

\bibliographystyle{splncs04}
\bibliography{main}

\clearpage
\appendix
\begin{center}
    \LARGE \bfseries Supplementary Materials
\end{center}
\vspace{1em} 

\section{Implementation Details for FutureVLA}
In this appendix, we provide a detailed description of the architecture and training configuration of FutureVLA, along with additional experiments that further validate the effectiveness of our method.

\subsection{Architecture Overview}

\noindent\textbf{Visual Tokenization.}
The pretraining model takes 17 consecutive RGB frames as input, with each frame resized to $224 \times 224$. To efficiently encode future-aware temporal information, we adopt the encoder of a pretrained 3D-VAE from WAN2.2\cite{wan2025wan} and keep it frozen during training.
Due to architectural constraints of the 3D-VAE, the number of input frames must follow the form $4N+1$. The input video is encoded via temporal block encoding, producing a compact temporal representation. For a 17-frames clip, the resulting video tokens have a resolution of $1960 \times 48$.

\noindent\textbf{Joint Visuomotor Gating.} The temporal video tokens are first processed by a lightweight encoder consisting of two Transformer layers. The encoded tokens are then evenly divided into two subsets with decoupled objectives: 980 visual tokens and 980 motor tokens.
The visual tokens are refined by a three-layer Transformer. Meanwhile, the motor tokens are processed by a single Transformer and subsequently attend to the visual tokens through cross-attention, where motor tokens act as queries and visual tokens serve as keys and values. To prevent visual domination, we introduce a gated update mechanism with a learnable scalar coefficient. The gated, visually-conditioned embeddings are adaptively weighted and added to the motor tokens, yielding the final joint visuomotor embeddings. This conditioning-and-update process is iteratively applied three times.

\noindent\textbf{Visual Reconstruction.}
The visual branch provides explicit supervision to the visual tokens by reconstructing the latent embedding of the first frame encoded by the 3D-VAE\cite{wan2025wan}, which has a resolution of $392 \times 48$. To align the number of visual tokens with the reconstruction target, we introduce a QueryPool module. Specifically, a set of learnable queries attends to the visual tokens to produce attention weights via a softmax operation, which are then used to aggregate visual tokens into a fixed number of pooled tokens. The pooled representations are finally fed into a three-layer Transformer decoder to predict the reconstructed latent embedding.

\subsection{Training Details}
\begin{figure}[t]
  \centering
   \includegraphics[width=0.6\linewidth]{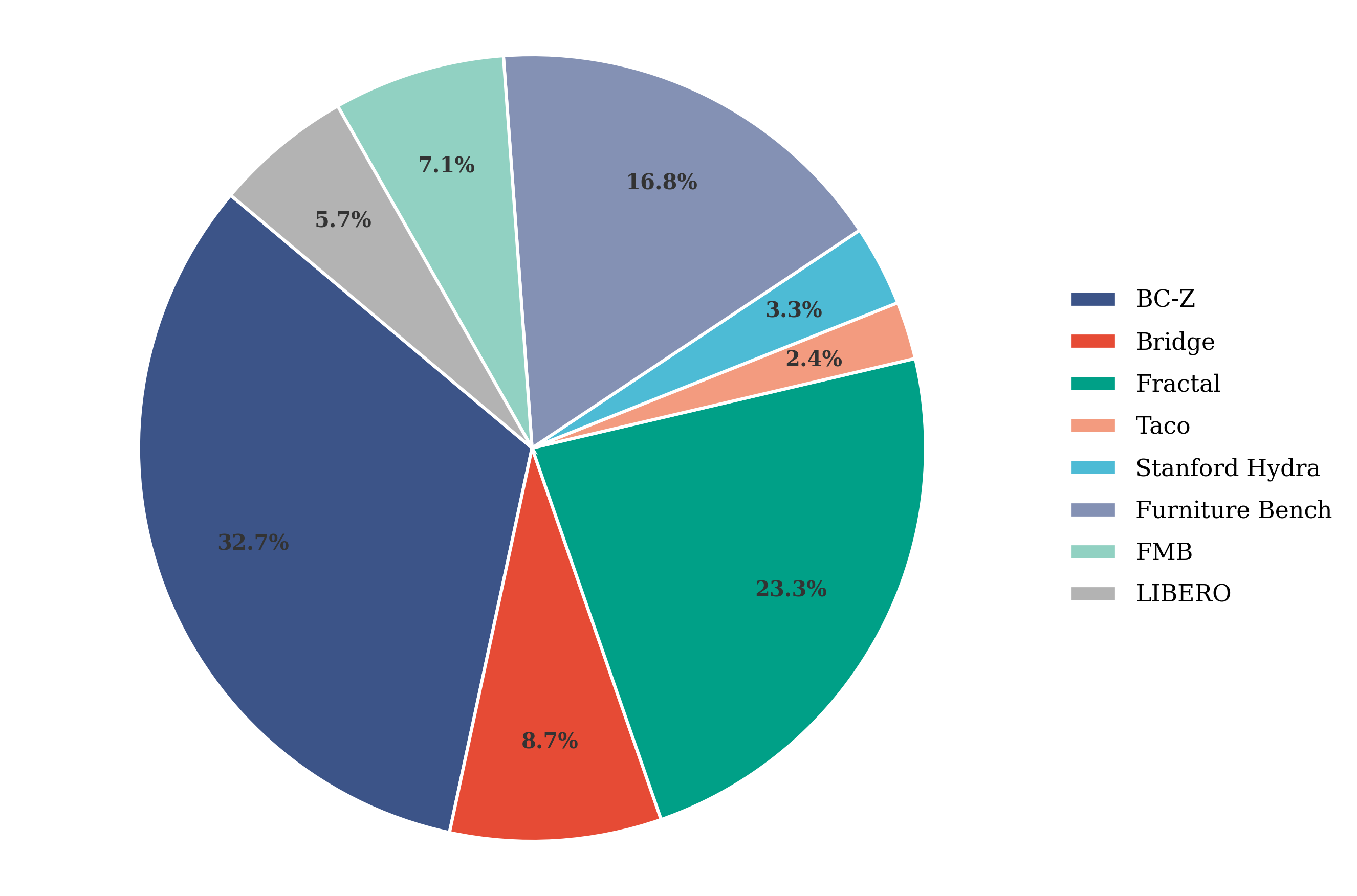}
   \caption{\small The composition of datasets used for Joint Visuomotor Pretraining.}
   \label{fig10}
\end{figure}

\noindent\textbf{Datasets.}
During pretraining stage, our model is trained on a mixture of datasets drawn primarily from OXE\cite{open_x_embodiment} and LIBERO\cite{Libero}, which comprises 15.6 million frames. The composition of the training data are illustrated in \cref{fig10}. The goal of this stage is to extract highly generalizable physical dynamics across diverse embodiments and environments, rather than overfitting to a specific task. For single-view datasets, the same image is duplicated to standardize inputs to a two-view format. Following common practice in VLA training, we use an action chunk size of 16 throughout pretraining.

\noindent\textbf{Training Protocol.}
We pretrain our model with a batch size of 256 and a base learning rate of $1\times10^{-5}$, using a linear warm-up schedule over the first 5{,}000 steps. Pretraining takes approximately three days on 16 NVIDIA A100 GPUs. For the post-training stage, we adopt Qwen3-VL-4B-Instruct~\cite{yang2025qwen3} as the VLM backbone and initialize it with pretrained weights. For the GR00T-style and OFT-style paradigms, we implement them according to their respective interaction mechanisms between the VLA module and the action head. The post-training stage also uses a 5{,}000-step linear warm-up schedule. To facilitate stable adaptation to downstream tasks, we apply a cosine decay schedule to the latent alignment loss coefficient $\beta$, progressively reducing the strength of latent supervision throughout training.

\subsection{Ablation Study Details}
We provide detailed ablation settings and additional experiments to further validate the effectiveness of our method.

\begin{table}[!t]
\centering
\caption{Comparison of different VLA models on the LIBERO-Plus benchmark. For simplicity, we abbreviate Camera/Robot/Language/Background as Cam./Rob./Lang./Back., respectively. ``GT'' indicates GR00T-style model, ``OT'' denotes OFT-style model.}
\setlength{\tabcolsep}{1.0mm}{
\begin{tabular}{ccccccccc}
\toprule
Methods & Cam. & Rob. & Lang. & Light & Back. & Noise & Layout & Total \\
\midrule
OpenVLA\cite{openvla} 
& 0.8 & 3.5 & 23.0 & 8.1 & 34.8 & 15.2 & 28.5 & 15.6 \\
OpenVLA-OFT\cite{openvla-oft} 
& 56.4 & 31.9 & 79.5 & 88.7 & 93.3 & 75.8 & 74.2 & 69.6 \\
$\pi_0$\cite{black2024pi_0}
& 13.8 & 6.0 & 58.8 & 85.0 & 81.4 & \textbf{79.0} & 68.9 & 53.6 \\
WorldVLA\cite{worldvla} 
& 0.1 & 27.9 & 41.6 & 43.7 & 17.1 & 10.9 & 38.0 & 25.0 \\
UniVLA\cite{bu2025univla} 
& 1.8 & 46.2 & 69.6 & 69.0 & 81.0 & 21.2 & 31.9 & 43.9 \\
\cellcolor{lightgray}FutureVLA-GT
& \cellcolor{lightgray}\textbf{59.7} & \cellcolor{lightgray}66.0 & \cellcolor{lightgray}88.2 
& \cellcolor{lightgray}97.4 & \cellcolor{lightgray}\textbf{97.8} & \cellcolor{lightgray}77.3 
& \cellcolor{lightgray}82.6 & \cellcolor{lightgray}\textbf{79.7} \\
\cellcolor{lightgray}FutureVLA-OT 
& \cellcolor{lightgray}50.4 & \cellcolor{lightgray}\textbf{68.8} & \cellcolor{lightgray}\textbf{90.0} 
& \cellcolor{lightgray}\textbf{97.9} & \cellcolor{lightgray}95.4 & \cellcolor{lightgray}72.6 
& \cellcolor{lightgray}\textbf{83.9} & \cellcolor{lightgray}78.2 \\
\bottomrule
\end{tabular}
}
\label{Tab.plus_main}
\end{table}

\noindent\textbf{Simulated Evaluation on LIBERO-Plus.}
To further evaluate the generalization and robustness of our method, we conduct experiments on the LIBERO-Plus\cite{fei2025libero} benchmark, which introduces diverse perturbations to the original evaluation setting. We directly adopt the training weights obtained on LIBERO\cite{Libero} for evaluation, and the detailed results are reported in \cref{Tab.plus_main}. Compared to OpenVLA-OFT\cite{openvla-oft}, our FutureVLA-GT achieves an absolute average gain of 10.1\% in overall success rate. Moreover, compared with the explicit future guidance method WorldVLA\cite{worldvla} and the implicit future guidance method UniVLA\cite{bu2025univla}, our approach improves the success rate by 54.7\% and 35.8\%, respectively. These results support our hypothesis that previous future guidance methods suffer from visual dominance, allocating excessive capacity to task-irrelevant appearance variations rather than robust control dynamics. In contrast, our method focuses on capturing true motor intent, enabling FutureVLA to maintain strong performance under diverse perturbations and exhibit superior generalization across a wide range of physical and semantic shifts.

\noindent\textbf{More Ablation of Impact of JVPM.} 
To comprehensively validate the effectiveness and generality of our Joint Visuomotor Predictive Modeling guidance across diverse environments and robot embodiments, we provide extended ablation studies comparing policies trained with (w/ JVPM) and without (wo/ JVPM) our latent embeddings alignment strategy on the Google robot, LIBERO benchmark, LIBERO-Plus benchmark, and real-world robot.

\begin{table}[!t]
\centering
\caption{Ablation of JVPM guidance on the Google robot.}
\setlength{\tabcolsep}{0.2mm}
\begin{tabular}{ccccccc}
\toprule
Settings & Methods
& \makecell{\small Pick\\Coke Can}
& \makecell{\small Move\\Near}
& \makecell{\small Open/Close\\Drawer}
& \makecell{\small Put in\\Drawer} & Avg. \\
\midrule
\multirow{4}{*}{\makecell{\small Visual\\Matching}}
& FutureVLA-GT wo/ JVPM
& 97.6 & 69.1 & 66.2 & 60.1 & 73.2 \\
& FutureVLA-GT w/ JVPM
& 92.3 & 74.2 & \textbf{68.5} & \textbf{85.2} & \textbf{80.1} \\
& FutureVLA-OT wo/ JVPM
& 93.0 & 70.4 & 54.2 & 29.6 & 61.8 \\
& FutureVLA-OT w/ JVPM
& \textbf{97.0} & \textbf{83.8} & 55.6 & 74.1 & 77.6 \\
\midrule
\multirow{4}{*}{\makecell{\small Variant\\Aggregation}}
& FutureVLA-GT wo/ JVPM
& 92.7 & 68.1 & \textbf{61.9} & 62.4 & 71.2 \\
& FutureVLA-GT w/ JVPM
& \textbf{95.7} & \textbf{72.9} & 52.4 & \textbf{81.2} & \textbf{75.6} \\
& FutureVLA-OT wo/ JVPM
& 50.9 & 23.5 & 40.2 & 7.9 & 30.6 \\
& FutureVLA-OT w/ JVPM
& 80.5 &61.5 & 31.7 & 23.8 & 49.4 \\
\bottomrule
\end{tabular}
\label{Tab.8}
\end{table}

\begin{table}[t]
    \centering
    \caption{Ablation of JVPM guidance on the LIBERO.}
    \setlength{\tabcolsep}{2mm}
    \begin{tabular}{cccccc}
        \toprule
        Methods & Object & Spatial & Goal & Long & Avg. \\
        \midrule
        FutureVLA-GT wo/JVPM
        & 98.6
        & 98.0
        & 97.0
        & 92.0
        & 96.4 \\

        FutureVLA-GT w/JVPM
        & \textbf{99.8}
        & \textbf{98.8}
        & \textbf{98.6}
        & \textbf{96.0}
        & \textbf{98.3} \\
\midrule
        FutureVLA-OT wo/JVPM
        & 98.4
        & 98.0
        & 96.6
        & 93.6
        & 96.6 \\

        FutureVLA-OT w/JVPM
        & \textbf{99.2}
        & \textbf{99.6}
        & \textbf{99.2}
        & \textbf{94.8}
        & \textbf{98.2} \\

        \bottomrule
    \end{tabular}
\label{Tab.9}
\end{table}

\begin{table}[!t]
\centering
\caption{Ablation of JVPM guidance on the LIBERO-Plus.}
\setlength{\tabcolsep}{1.0mm}{
\begin{tabular}{ccccccccc}
\toprule
Methods & Cam. & Rob. & Lang. & Light & Back. & Noise & Layout & Total \\
\midrule
FutureVLA-GT wo/ JVPM 
& 48.8 & 60.3 & 87.1 & 95.5 & 95.2 & 75.1 & 78.3 & 75.4 \\
FutureVLA-GT w/ JVPM 
& \textbf{59.7} & \textbf{66.0} & \textbf{88.2} & \textbf{97.4} & \textbf{97.8} & \textbf{77.3} & \textbf{82.6} & \textbf{79.7} \\
\midrule
FutureVLA-OT wo/ JVPM 
& 47.1 & 59.8 & 87.3 & 95.9 & 93.9 & \textbf{73.2} & 78.9 & 74.8 \\
FutureVLA-OT w/ JVPM 
& \textbf{50.4} & \textbf{68.8} & \textbf{90.0} & \textbf{97.9} & \textbf{95.4} & 72.6 & \textbf{83.9} & \textbf{78.2} \\
\bottomrule
\end{tabular}
}
\label{Tab.plus_ablation}
\end{table}

\begin{table}[!t]
    \centering
    \caption{Ablation of JVPM guidance on real-world robotic tasks.}
    \setlength{\tabcolsep}{2mm}
    \begin{tabular}{cccccc}
        \toprule
        Methods 
        & \makecell{\small Make \\ Burger}
        & \makecell{\small Insert \\ Rose}
        & \makecell{\small Scoop \\ Beans}
        & \makecell{\small Eraser \\ Handwriting}
         & Avg. \\
        \midrule
        FutureVLA-GT wo/ JVPM & 80.0 & 33.3 & 46.7 & 33.3 & 48.3 \\
        FutureVLA-GT w/ JVPM
        & \textbf{100.0} & \textbf{46.7} &\textbf{60.0} & \textbf{73.3} & \textbf{70.0} \\
        \midrule
        FutureVLA-OT wo/ JVPM & 40.0 & 26.7 & 40.0 & 26.7 & 33.3 \\
        FutureVLA-OT w/ JVPM
        & \textbf{60.0} & \textbf{40.0} & \textbf{60.0} & \textbf{46.7} & \textbf{51.7} \\
        \bottomrule
    \end{tabular}
\label{Tab.10}
\end{table}

As shown in \cref{Tab.8}, injecting joint visuomotor priors brings robust improvements. In terms of average success rate, JVPM yields absolute gains of 6.9\% and 4.4\% for the GR00T-style model under the Visual Matching and Variant Aggregation settings, respectively. The OFT-style variant exhibits even greater sensitivity to our future-aware guidance, achieving remarkable improvements of 15.8\% and 18.8\% under the same settings.

As for the multi-task LIBERO benchmark, introducing our joint visuomotor embeddings also leads to consistent gains. As shown in \cref{Tab.9}, the GR00T-style and OFT-style frameworks attain average success rate improvements of 1.9\% and 1.6\%, respectively. Specifically, the GR00T-style variant achieves a 4\% absolute gain on the challenging \textit{Long} tasks suite, corroborating our claim that temporally continuous embeddings provide critical guidance for extended action horizons.

As detailed in \cref{Tab.plus_ablation}, removing the JVPM module leads to noticeable performance degradation across both GR00T-style and OFT-style frameworks. Specifically, in the FutureVLA-OT formulation, the absence of JVPM reduces the success rate under \textit{Robot} initializations from 68.8\% to 59.8\%, and under \textit{Layout} shifts from 83.9\% to 78.9\%. By employing a gated cross-attention mechanism, JVPM enables the motor stream to selectively query spatial affordances from visual tokens, allowing it to focus on continuous physical dynamics. This design prevents the embeddings from collapsing into mere visual residuals, thereby equipping the downstream VLA with a more robust and physics-aware internal model.

As illustrated in \cref{Tab.10}, across four real-world contact-rich tasks, incorporating joint visuomotor priors improves the average success rate by an impressive 21.7\% and 18.4\% for the GR00T-style and OFT-style models, respectively. In the challenging whiteboard erasing task, which demands sustained and stable force regulation, the success rate surges by 40.0\% for FutureVLA-GT and 20.0\% for FutureVLA-OT when JVPM guidance is applied. This confirms that our pretraining stage captures highly effective physical dynamics that translate perfectly into real-world control enhancements.

\begin{table}[t]
    \centering
    \caption{Impact of the reconstruction loss weight $\lambda$ on the WidowX robot.}
    \setlength{\tabcolsep}{1.5mm}
    \begin{tabular}{c|ccccc}
        \toprule
        $\lambda$ & \makecell{\small Put \\ Spoon}
        & \makecell{\small Put \\Carrot}
        & \makecell{\small Stack \\Cube}
        & \makecell{\small Put \\Eggplant} & Avg. \\
        \midrule
        0.5 & 70.8 & 75.0 & 29.2 & 95.8 & 67.7 \\
        1.0 & \textbf{83.3} & \textbf{75.0} & \textbf{29.2} & \textbf{100.0} & \textbf{71.9} \\
        1.5 & 79.2 & 70.8 & 29.2 & 100.0 & 69.8 \\
        \bottomrule
    \end{tabular}
\label{Tab.11}
\end{table}

\begin{table}[t]
    \centering
    \caption{Effect of different video encoding strategies on performance and computational efficiency. ``Cost time'' denotes the average training time per step.}
    \setlength{\tabcolsep}{1.mm}
    \begin{tabular}{cc|c|ccccc}
        \toprule
        VQ-GAN & 3D-VAE & Cost time (s) & \makecell{\small Put \\ Spoon}
        & \makecell{\small Put \\Carrot}
        & \makecell{\small Stack \\Cube}
        & \makecell{\small Put \\Eggplant} & Avg. \\
        \midrule
        \checkmark &  & 3.53 & 79.2 & 62.5 & 20.8 & 95.8 & 64.6 \\
                  & \checkmark & 3.28 & \textbf{83.3} & \textbf{75.0} & \textbf{29.2} & \textbf{100.0} & \textbf{71.9} \\
        \bottomrule
    \end{tabular}
\label{Tab.12}
\end{table}

\begin{table}[t]
    \centering
    \caption{Impact of different reconstruction targets on the WidowX robot.}
    \label{tab:frame_comparison}
    \setlength{\tabcolsep}{5pt}
    \begin{tabular}{cccccc}
        \toprule
        Reconstruction Target& \makecell{\small Put \\ Spoon}
        & \makecell{\small Put \\Carrot}
        & \makecell{\small Stack \\Cube}
        & \makecell{\small Put \\Eggplant} & Avg. \\
        \midrule
        Last-frame  & 75.0 & \textbf{75.0} & \textbf{29.2} & 95.8 & 68.8 \\
        First-frame & \textbf{83.3} & \textbf{75.0} & \textbf{29.2} & \textbf{100.0} & \textbf{71.9} \\
        \bottomrule
    \end{tabular}
\label{Tab.13}
\end{table}

\begin{figure}[t]
  \centering
   \includegraphics[width=0.8\linewidth]{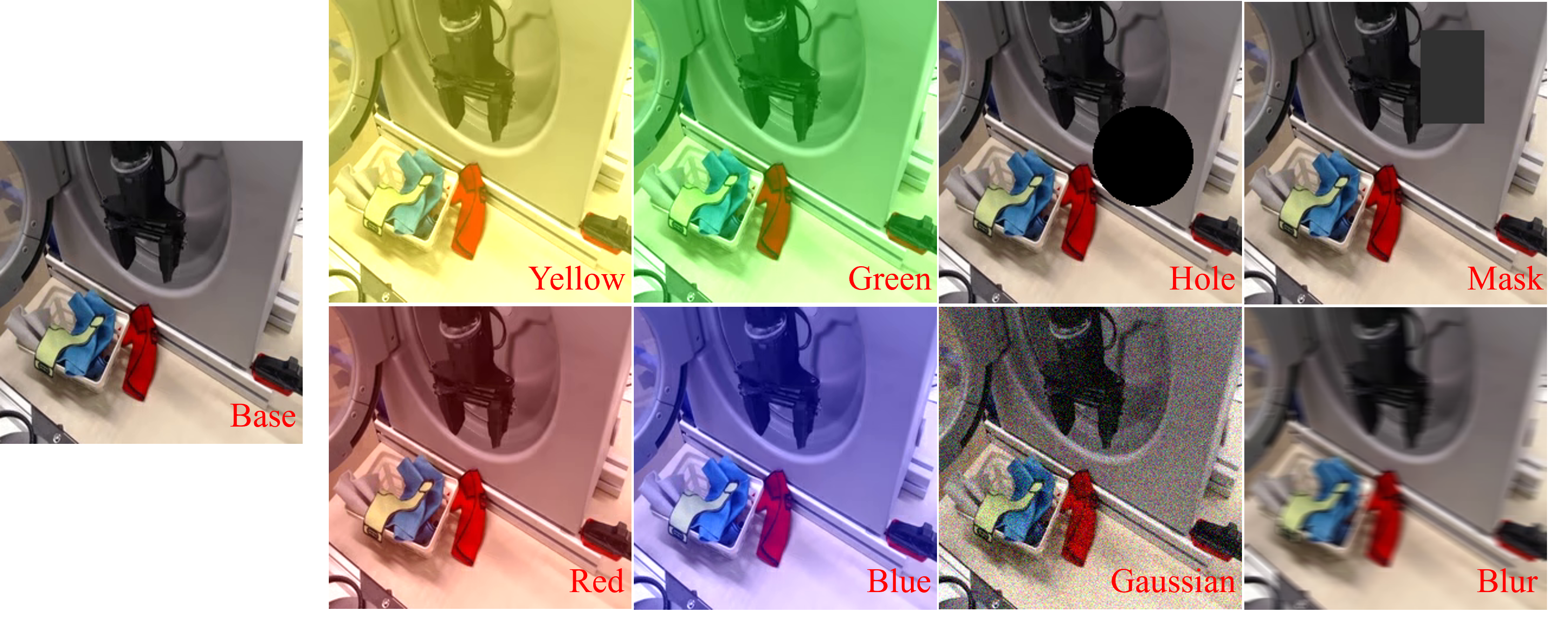}
   \caption{\small Visualization of visual perturbations applied to video frames. The left shows the original frame, and the right illustrates different types of data perturbations.}
   \label{fig11}
\end{figure}

\begin{table}[t]
\centering
\caption{Performance comparison of LAPA-style, Villa-X-style, and our method under visual perturbations. 
``Embedding MSE'' measures the discrepancy between latent embeddings before and after noise injection.
``Origin Action MSE'' and ``Noise Action MSE'' denote the error between predicted actions and ground-truth actions under clean and perturbed observations, respectively.
``Action Bias'' quantifies the action deviation induced by visual perturbations. }
\label{tab:performance_comparison}
\begin{tabular}{cccc}
\toprule
Methods & LAPA-style & Villa-X-style & \textbf{Ours} \\
\midrule
Embedding MSE     & 0.3047 & 0.0188 & \textbf{0.0054} \\
Origin Action MSE & 0.0691 & 0.0399 & \textbf{0.0371} \\
Noise Action MSE  & 0.0752 & 0.0435 & \textbf{0.0398} \\
Action Bias        & 0.0061 & 0.0036 & \textbf{0.0027} \\
\bottomrule
\end{tabular}
\label{Tab.14}
\end{table}


\begin{table}[t]
    \centering
    \caption{Frame index selection strategy under different frame numbers.}
    \label{tab:frame_index}
    \setlength{\tabcolsep}{6pt}
    \begin{tabular}{cc}
        \toprule
        Frame number & Selected frame indices \\
        \midrule
        2  & 1, 1, 17, 17, 17 \\
        5  & 1, 6, 10, 14, 17 \\
        9  & 1, 3, 5, 7, 9, 11, 13, 15, 17 \\
        17 & 1--17 \\
        \bottomrule
    \end{tabular}
    \label{Tab.16}
\end{table}

\noindent\textbf{Impact of Ratio in Reconstruction Loss.} Here, we conduct a coefficient ablation study on the reconstruction loss during pretraining, as shown in \cref{Tab.11}. We observe that both overly large and overly small reconstruction loss weights lead to performance degradation. This indicates that an appropriate balance between reconstruction and action supervision is crucial for effective training. Based on this analysis, we set the coefficients of both the reconstruction loss and the action loss to 1 in all experiments.

\noindent\textbf{Effectiveness of Joint Temporal Encoding.}
We further evaluate the efficiency and coherence of our temporal embeddings extraction. We compare a single-frame encoding baseline, where each frame is independently processed by a pretrained VQ-GAN~\cite{esser2021taming} and concatenated into temporal tokens, with our proposed multi-frame 3D-VAE encoding. As reported in~\cref{Tab.12}, the 3D-VAE consistently achieves higher task success rates while significantly reducing per-step training time. This indicates that joint temporal encoding explicitly captures inter-frame motion dynamics, yielding more coherent temporal representations and improved computational efficiency for downstream VLA models.

\noindent\textbf{Impact of the Reconstruction Target.} We investigate the rationale behind our decoupled supervision design by replacing the default first-frame reconstruction with last-frame reconstruction. As shown in \cref{Tab.13}, reconstructing the last frame results in a consistent performance degradation of 3.1\% on average across all metrics compared to reconstructing the first frame.
This phenomenon perfectly aligns with our theoretical claims. Reconstructing the first frame serves as a static geometric anchor, supplying stable initial scene constraints  that condition the motor stream. In contrast, forcing the visual tokens to reconstruct the last frame requires them to memorize future visual dynamics. This re-introduces the visually-dominated embedding entanglement seen in implicit baseline methods, weakening the purity of the physical motor modeling. Thus, first-frame reconstruction is strictly necessary to achieve true visually-conditioned supervision decoupling.

\noindent\textbf{Robustness against Visually-Dominated Entanglement.}
To further validate that our decoupled architecture successfully prevents visual domination, we evaluate embedding robustness under task-irrelevant visual perturbations (e.g., color jittering, Gaussian noise) as illustrated in~\cref{fig11}. As reported in~\cref{Tab.14}, because LAPA and Villa-X entangle visual reconstruction with latent action learning, introducing visual noise causes embedding deviation (high Embedding MSE) and consequent action execution bias. Conversely, FutureVLA inherently isolates visual rendering from motor execution. By structurally decoupling visual state preservation from the motor stream, our Joint Visuomotor Gating mechanism effectively shields the motor tokens from being corrupted by task-irrelevant visual dynamics. This architectural advantage yields substantially lower embedding deviation and robust action prediction under visually noisy conditions.


\noindent\textbf{Details of Necessity of Continuous Temporal Alignment.}
This section details the experimental setup for the contribution of inter-frame temporal information. We start from a sequence of 17 consecutive frames as the full temporal context and examine the effect of temporal sparsity by uniformly sampling subsets from this sequence. We evaluate four sampling configurations with sizes $\{2, 5, 9, 17\}$. Due to architectural constraints of the 3D-VAE in WAN\cite{wan2025wan}, which requires the number of input frames to follow a $4N+1$ format, the two-frame setting is padded to five frames via frame repetition to ensure compatibility. The exact frame indices used for each configuration are summarized in \cref{Tab.16}.

\noindent\textbf{Physics-Aware Action Consistency Score.}
\label{pasc}
A naive cosine similarity between action vectors is insufficient for evaluating action chunk similarity, as it ignores the magnitude of control signals and assumes strict temporal alignment between action sequences. These assumptions are often violated in real robotic executions, where temporal variations and execution speed differences naturally arise. To address these limitations, we propose a Physics-Aware Action Consistency (PAAC) metric.

Our metric builds upon Dynamic Time Warping (DTW) to explicitly account for temporal misalignment while preserving sensitivity to the physical magnitude of control signals. Let $\mathcal{A} = \{ A_1, \dots, A_N \}$ denote a set of action sequences, where each $A_i \in \mathbb{R}^{T \times D}$ consists of $T$ time steps and $D$ actuation dimensions. For a pair of action sequences $A_i$ and $A_j$, we define their physical discrepancy using the DTW distance:
\begin{equation}
    \mathcal{D}_{\text{DTW}}(A_i, A_j) =
    \min_{\pi \in \Pi}
    \sum_{(k, m) \in \pi}
    \left\| A_{i,k} - A_{j,m} \right\|_2^2,
\end{equation}
where $\Pi$ denotes the set of all valid monotonic alignment paths between the two sequences, and $\|\cdot\|_2$ is the Euclidean norm, ensuring sensitivity to the physical magnitude of actions.

Since the DTW distance is unbounded, we convert it into a normalized similarity score $S_{\text{phys}} \in [0,1]$ using a Radial Basis Function (RBF) kernel:
\begin{equation}
    S_{\text{phys}}(A_i, A_j) =
    \exp\left(
        - \frac{\mathcal{D}_{\text{DTW}}(A_i, A_j)}{2\sigma^2}
    \right).
\end{equation}

The bandwidth parameter $\sigma$ controls the sensitivity of the similarity measure. To ensure robustness across different action distributions, we adopt the median heuristic, setting $\sigma$ to the median of all pairwise DTW distances:
\begin{equation}
    \sigma =
    \mathrm{median}
    \left(
        \left\{
            \sqrt{\mathcal{D}_{\text{DTW}}(A_i, A_j)}
            \mid i \neq j
        \right\}
    \right).
\end{equation}

The proposed PAAC metric assigns high similarity scores only to action chunks that are both temporally consistent and physically coherent. It therefore provides a meaningful and stringent ground-truth signal for evaluating the alignment between latent embeddings and executed robot behaviors. The complete computation procedure is summarized in \cref{alg:paac}. Based on this metric, we compute an action similarity matrix and correlate it with the latent action similarity matrix to obtain the probability density distributions shown in Fig.8 in main paper.

\begin{algorithm}[!t]
\caption{Physics-Aware Action Consistency}
\label{alg:paac}
\KwIn{Action sequences $\mathcal{A} = \{A_1, \dots, A_N\}$.}

\For{$i, j \in \{1, \dots, N\}$}{
    $\text{cost}_{m,n} = \|A_{i,m} - A_{j,n}\|_2^2 \quad \forall m, n \in \{1, \dots, T\}$ \\
    $\mathcal{D}_{ij} \leftarrow \text{DTW}(A_i, A_j, \text{cost})$ 
}

$\sigma \leftarrow \text{median}\left( \{ \sqrt{\mathcal{D}_{ij}} \mid i \neq j \} \right)$ 

\For{$i, j \in \{1, \dots, N\}$}{
    $S_{phys}(i, j) = \exp\left( - \frac{\mathcal{D}_{ij}}{2\sigma^2} \right)$ 
}

\Return{Action Similarity matrix $\mathcal{S}_{phys}$}\;
\end{algorithm}

\subsection{Limitations and Future Work}
Despite its promising results, FutureVLA has limitations that present exciting avenues for future research. For contact-rich operations, such as erasing a whiteboard, relying strictly on visual constraints can be insufficient. Future iterations of our joint visuomotor predictive architecture would greatly benefit from integrating multi-sensory feedback, such as tactile or force-torque signals, to ground the predictive modeling in a more comprehensive physical reality.

\subsection{Simulation Evaluation Details}
We provide detailed descriptions of the simulation benchmarks used in our experiments.

\noindent\textbf{SimplerEnv Benchmark.}
To evaluate cross-embodiment generalization and the correlation between simulation and real-world performance, we conduct experiments on the SimplerEnv benchmark\cite{simpleenv} with two robot embodiments: the Google robot and the WidowX robot.

For the Google robot, we consider two evaluation settings. \textit{Visual Matching} evaluates spatial generalization under variations in camera viewpoints and object placements, while \textit{Variant Aggregation} assesses zero-shot robustness to environmental perturbations such as changes in lighting conditions and background textures. Both settings include four representative manipulation tasks: \textit{Pick Coke Can}, \textit{Move Near}, \textit{Open/Close Drawer}, and \textit{Put in Drawer}.

For the WidowX robot, we adopt the \textit{Visual Matching} setting to align with the data distribution of the Bridge dataset\cite{Bridge_data}. Evaluation is performed on four manipulation tasks: \textit{Put Carrot on Plate} (\textit{Put Carrot}), \textit{Put Spoon on Towel} (\textit{Put Spoon}), \textit{Stack Green on Yellow} (\textit{Stack Cube}), and \textit{Put Eggplant in Basket} (\textit{Put Eggplant}). All SimplerEnv experiments follow the standard evaluation protocols defined in CogAct\cite{cogact}.

\noindent\textbf{LIBERO Benchmark.}
We further evaluate our method on the LIBERO benchmark\cite{Libero} to analyze different aspects of temporal reasoning and knowledge transfer. Specifically, we consider four task suites: \textit{Spatial}, which evaluates spatial reasoning under varying object layouts; \textit{Object}, which focuses on object-centric manipulation across diverse object categories; \textit{Goal}, which tests goal-conditioned instruction following in static environments; and \textit{Long}, which emphasizes long-horizon planning across multi-stage tasks. Each task suite contains 10 tasks, with 50 human-teleoperated demonstrations provided per task.

\noindent\textbf{LIBERO-Plus Benchmark.}
We further evaluate FutureVLA on the LIBERO-Plus benchmark\cite{fei2025libero} to systematically analyze its robustness and generalization capabilities under diverse environmental variations. Building upon the core scenarios of the original dataset, LIBERO-Plus introduces controlled perturbations across seven distinct dimensions: \textit{Objects Layout}, \textit{Camera Viewpoints}, \textit{Robot Initial States}, \textit{Language Instructions}, \textit{Light Conditions}, \textit{Background Textures}, and \textit{Sensor Noise}. To rigorously test the limits of our joint visuomotor modeling, this comprehensive benchmark comprises a total of 10,030 tasks, which are further stratified into five fine-grained difficulty levels, providing a thorough diagnostic framework for real-world visuomotor reliability.

\subsection{Real-World Robot Evaluation Details}
We provide detailed descriptions of the real-world robotic platform and task suites used in our experiments.

\noindent\textbf{Real-World Robot Setup.}
All real-world experiments are conducted on a Franka Research 3 robotic platform, which is widely used in robotics and embodied AI research. The system features a 7-DoF manipulator with joint-level torque and force sensing. For object manipulation, we employ a 1-DoF Panda Hand gripper. Visual observations are captured using a single Intel RealSense D435i RGB-D camera mounted in an eye-on-base configuration, providing a global view of the workspace. In addition, we employ another camera mounted on the wrist to capture close-up observations during manipulation. The robot is equipped with a Robotiq 2F-85 gripper.

\noindent\textbf{Baseline Models.} (1) OpenVLA~\cite{openvla}: Initialized from the publicly released checkpoints, which is pretrained on OXE~\cite{open_x_embodiment}, and we extended the model to process both third-person and wrist-view images. (2) $\pi_0$~\cite{black2024pi_0}:
Fine-tuned from the official JAX checkpoints of the Droid~\cite{khazatsky2024droid} dataset.

\noindent\textbf{Task Suites.}
We evaluate our method on four real-world manipulation tasks that span diverse challenges in sequential reasoning, tool use, contact-rich interaction, and high-precision insertion:
\begin{itemize}
    \item \textbf{Make a Burger.} 
    This task evaluates multi-stage sequential assembly and spatial reasoning. The robot is required to grasp and stack multiple ingredients in a predefined semantic order while maintaining physical stability throughout the assembly process. To promote generalization, the initial positions of bread slices and meat are randomized across different shelf locations, requiring accurate visual perception and closed-loop control.

    \item \textbf{Insert Roses into a Pot.}
    This task evaluates high-precision insertion under narrow spatial constraints. The robot must extract flowers from a source pot and insert them into a target pot. Due to the slender geometry of the stems and the limited clearance of the target openings, the task requires accurate spatial localization and fine-grained end-effector adjustments.

    \item \textbf{Scoop Beans with a Spoon.} 
    This task assesses tool use and manipulation of granular materials with complex dynamics. The robot must grasp a spoon, execute a coordinated scooping action to transfer beans from one container to another, maintain an appropriate end-effector orientation to prevent spillage, and finally place the spoon down. The task demands precise 6-DoF control and coherent long-horizon action execution.

    \item \textbf{Erase Handwriting on a Whiteboard.}
    This task focuses on contact-rich manipulation and continuous surface interaction. The robot must grasp an eraser and maintain stable contact with a vertical whiteboard while performing repeated wiping actions. Successful completion requires visual closed-loop control to adaptively cover remaining handwritten regions, testing fine-grained perception and reactive action planning.

\end{itemize}

For real-world post-training, we collect 300 trajectories in total, with 75 trajectories per task, recorded at 5~Hz. We post-process the collected data to remove static frames with zero-action signals.

\subsection{Details of Compared Models}
We compare our method with representative VLA approaches that span three categories: original VLA baselines, explicit future guidance methods, and implicit latent guidance methods.

\noindent\textbf{Original Baselines.}
We first compare against several widely adopted VLA baselines that directly map multimodal observations to robot actions.

\begin{itemize}
    \item \textbf{OpenVLA-OFT\cite{openvla-oft}.}
    OpenVLA-OFT introduces an optimized post-training strategy to efficiently adapt vlm to robotic control. It combines parallel decoding, action chunking, continuous action representations, and an regression objective to improve both inference efficiency and performance. For tasks requiring stronger language grounding, FiLM conditioning is used to inject linguistic information into action generation.

    \item \textbf{$\pi_0$\cite{black2024pi_0}.}
    $\pi_0$ is a large-scale generalist VLA developed by Physical Intelligence. It employs a flow-matching objective to generate smooth, high-frequency action trajectories and is trained on a diverse dataset spanning seven robotic platforms and 68 tasks. This design enables strong zero-shot generalization and robustness in complex multi-stage manipulation tasks.

    \item \textbf{GR00T-N1.5\cite{bjorck2025gr00t}.}
    GR00T-N1.5 is a foundation model for generalist humanoid robots that maps multimodal observations to continuous actions using a flow-matching transformer. It jointly encodes multi-view visual inputs and proprioceptive signals within a unified architecture, supporting robust loco-manipulation across embodiments. Pre-trained on large-scale datasets, GR00T-N1.5 demonstrates strong scalability in simulation and transferability to real-world hardware through post-training.
\end{itemize}

\noindent\textbf{Explicit Future Guidance Methods.}
We further compare with methods that explicitly incorporate future information by predicting future observations or structured representations to guide action generation.

\begin{itemize}
    \item \textbf{WorldVLA\cite{worldvla}.}
    WorldVLA jointly models action generation and future observation prediction within an autoregressive action-world framework. By learning environment dynamics through video prediction, it augments action generation with world-model priors. To reduce error accumulation during rollout, WorldVLA employs selective attention masking that decouples historical action tokens from current predictions, improving long-horizon stability.

    \item \textbf{DreamVLA\cite{zhang2025dreamvla}.}
    DreamVLA formulates VLA learning as a perception-prediction-action loop by explicitly forecasting structured future representations, such as depth, dynamic regions, and semantic cues, rather than raw pixels. These predictions serve as intermediate guidance for action planning. The model adopts block-wise structured attention to prevent information leakage and uses a diffusion-based transformer for action decoding.
\end{itemize}

\noindent\textbf{Implicit Future Guidance Methods.}
Finally, we compare with approaches that provide future guidance implicitly by learning latent action representations.

\begin{itemize}
    \item \textbf{LAPA\cite{ye2025lapa}.}
    Latent Action Pretraining (LAPA) is an framework that enables VLAs to learn from large-scale videos. It uses a VQ-VAE to discretize inter-frame visual differences into latent action tokens that capture action semantics. A latent VLA is pretrained to predict these tokens and subsequently fine-tuned on robot data, enabling strong semantic abstraction and cross-embodiment transfer.

    \item \textbf{UniVLA\cite{bu2025univla}.}
    UniVLA proposes a unified cross-embodiment framework that derives task-centric latent actions from heterogeneous data sources. By separating task-independent and task-related tokens and applying phased exclusion and extraction, UniVLA models latent actions that capture action change characteristics relevant to downstream tasks.

    \item \textbf{Villa-X\cite{chen2025villa}.}
    Villa-X extends latent action modeling by jointly learning latent and robot action distributions within a unified diffusion framework. To ensure physical executability, it introduces a proprioceptive forward dynamics model as an auxiliary objective, grounding latent representations in feasible robot actions. 
\end{itemize}

%
%

\end{document}